\documentclass[acmtog]{acmart}
\usepackage{booktabs} 
\usepackage{graphicx}
\usepackage{amsmath}
\usepackage{booktabs}
\usepackage{multirow}
\usepackage{colortbl}
\usepackage{makecell}
\usepackage{diagbox}
\usepackage{xcolor}
\usepackage{pifont}%
\usepackage{threeparttable}
\usepackage[utf8]{inputenc} 
\usepackage[T1]{fontenc}    
\usepackage{graphicx}
\usepackage[ruled]{algorithm2e} 
\SetAlFnt{\small}
\SetAlCapFnt{\small}
\SetAlCapNameFnt{\small}
\SetAlCapHSkip{0pt}

\begin{document}
\title{Dyn-E: Local Appearance Editing of Dynamic Neural Radiance Fields}

\author{Shangzan Zhang}
\affiliation{%
 \institution{State Key Laboratory of CAD\&CG, Zhejiang University}
 \country{China}
 }
\author{Sida Peng}
\affiliation{%
 \institution{Zhejiang University}
 \country{China}
 }

\author{Yinji ShenTu}
\affiliation{%
 \institution{Zhejiang University}
 \country{China}
}
\author{Qing Shuai}
\affiliation{%
 \institution{State Key Laboratory of CAD\&CG, Zhejiang University}
 \country{China}
 }
\author{Tianrun Chen}
\affiliation{%
 \institution{Zhejiang University}
 \country{China}
}

\author{Kaicheng Yu}
\affiliation{%
 \institution{Alibaba Group}
 \country{China}
 }
\author{Hujun Bao}
\affiliation{%
 \institution{State Key Laboratory of CAD\&CG, Zhejiang University}
 \country{China}
 }
\author{Xiaowei Zhou}
\affiliation{%
 \institution{State Key Laboratory of CAD\&CG, Zhejiang University}
 \country{China}
 }
\authornote{Corresponding author.}





\begin{teaserfigure}
\centering
\includegraphics[width=\textwidth]{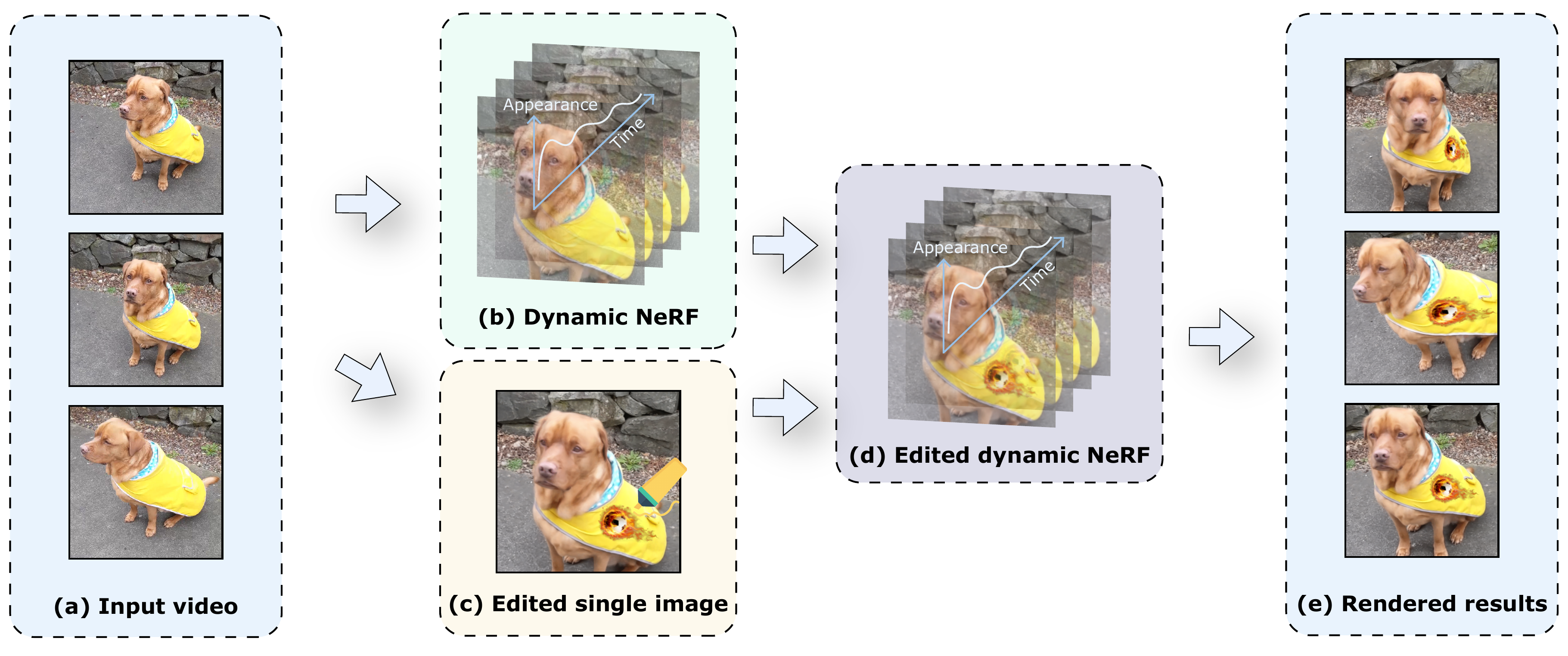}
\caption{Our proposed approach allows users to locally edit the appearance of a dynamic 3D 
scene in a user-friendly manner. Given training videos (a) and the reconstructed dynamic NeRF (b) as input, users can edit the appearance of the dynamic NeRF (d) by manipulating pixels in a single image (c). 
Experiment results demonstrate that our method can produce spatially and temporally consistent renderings (e). 
} 
\label{teaser} 
\end{teaserfigure}
    


%


\begin{abstract}
Recently, the editing of neural radiance fields (NeRFs) has gained considerable attention, but most prior works focus on static scenes while research on the appearance editing of dynamic scenes is relatively lacking.
In this paper, we propose a novel framework to edit the local appearance of dynamic NeRFs by manipulating pixels in a single frame of training video. 
Specifically, to locally edit the appearance of dynamic NeRFs while preserving unedited regions, we introduce a local surface representation of the edited region, which can be inserted into and rendered along with the original NeRF and warped to arbitrary other frames through a learned invertible motion representation network. By employing our method, users without professional expertise can easily add desired content to the appearance of a dynamic scene. We extensively evaluate our approach on various scenes and show that our approach achieves spatially and temporally consistent editing results. Notably, our approach is versatile and applicable to different variants of dynamic NeRF representations.
\end{abstract}


\makeatletter
\let\@authorsaddresses\@empty
\makeatother
\maketitle


\newcommand{\bK}{\mathbf{K}}
\newcommand{\bX}{\mathbf{X}}
\newcommand{\bY}{\mathbf{Y}}
\newcommand{\bk}{\mathbf{k}}
\newcommand{\bx}{\mathbf{x}}
\newcommand{\by}{\mathbf{y}}
\newcommand{\bhy}{\hat{\mathbf{y}}}
\newcommand{\bty}{\tilde{\mathbf{y}}}
\newcommand{\bG}{\mathbf{G}}
\newcommand{\bI}{\mathbf{I}}
\newcommand{\bg}{\mathbf{g}}
\newcommand{\bS}{\mathbf{S}}
\newcommand{\bs}{\mathbf{s}}
\newcommand{\bM}{\mathbf{M}}
\newcommand{\bw}{\mathbf{w}}
\newcommand{\eye}{\mathbf{I}}
\newcommand{\bU}{\mathbf{U}}
\newcommand{\bV}{\mathbf{V}}
\newcommand{\bW}{\mathbf{W}}
\newcommand{\bn}{\mathbf{n}}
\newcommand{\bv}{\mathbf{v}}
\newcommand{\bq}{\mathbf{q}}
\newcommand{\bR}{\mathbf{R}}
\newcommand{\bi}{\mathbf{i}}
\newcommand{\bj}{\mathbf{j}}
\newcommand{\bp}{\mathbf{p}}
\newcommand{\bt}{\mathbf{t}}
\newcommand{\bJ}{\mathbf{J}}
\newcommand{\bu}{\mathbf{u}}
\newcommand{\bB}{\mathbf{B}}
\newcommand{\bD}{\mathbf{D}}
\newcommand{\bz}{\mathbf{z}}
\newcommand{\bP}{\mathbf{P}}
\newcommand{\bC}{\mathbf{C}}
\newcommand{\bA}{\mathbf{A}}
\newcommand{\bZ}{\mathbf{Z}}
\newcommand{\bff}{\mathbf{f}}
\newcommand{\bF}{\mathbf{F}}
\newcommand{\bo}{\mathbf{o}}
\newcommand{\bO}{\mathbf{O}}
\newcommand{\bc}{\mathbf{c}}
\newcommand{\bm}{\mathbf{m}}
\newcommand{\bT}{\mathbf{T}}
\newcommand{\bQ}{\mathbf{Q}}
\newcommand{\bL}{\mathbf{L}}
\newcommand{\bl}{\mathbf{l}}
\newcommand{\ba}{\mathbf{a}}
\newcommand{\bE}{\mathbf{E}}
\newcommand{\bH}{\mathbf{H}}
\newcommand{\bd}{\mathbf{d}}
\newcommand{\br}{\mathbf{r}}
\newcommand{\be}{\mathbf{e}}
\newcommand{\bb}{\mathbf{b}}
\newcommand{\bh}{\mathbf{h}}
\newcommand{\bhh}{\hat{\mathbf{h}}}
\newcommand{\btheta}{\boldsymbol{\theta}}
\newcommand{\bTheta}{\boldsymbol{\Theta}}
\newcommand{\bpi}{\boldsymbol{\pi}}
\newcommand{\bphi}{\boldsymbol{\phi}}
\newcommand{\bPhi}{\boldsymbol{\Phi}}
\newcommand{\bmu}{\boldsymbol{\mu}}
\newcommand{\bSigma}{\boldsymbol{\Sigma}}
\newcommand{\bGamma}{\boldsymbol{\Gamma}}
\newcommand{\bbeta}{\boldsymbol{\beta}}
\newcommand{\bomega}{\boldsymbol{\omega}}
\newcommand{\blambda}{\boldsymbol{\lambda}}
\newcommand{\bLambda}{\boldsymbol{\Lambda}}
\newcommand{\bkappa}{\boldsymbol{\kappa}}
\newcommand{\btau}{\boldsymbol{\tau}}
\newcommand{\balpha}{\boldsymbol{\alpha}}
\newcommand{\nR}{\mathbb{R}}
\newcommand{\nN}{\mathbb{N}}
\newcommand{\nL}{\mathbb{L}}
\newcommand{\nF}{\mathbb{F}}
\newcommand{\nS}{\mathbb{S}}
\newcommand{\cN}{\mathcal{N}}
\newcommand{\cM}{\mathcal{M}}
\newcommand{\cR}{\mathcal{R}}
\newcommand{\cB}{\mathcal{B}}
\newcommand{\cL}{\mathcal{L}}
\newcommand{\cH}{\mathcal{H}}
\newcommand{\cS}{\mathcal{S}}
\newcommand{\cT}{\mathcal{T}}
\newcommand{\cO}{\mathcal{O}}
\newcommand{\cC}{\mathcal{C}}
\newcommand{\cP}{\mathcal{P}}
\newcommand{\cE}{\mathcal{E}}
\newcommand{\cF}{\mathcal{F}}
\newcommand{\cK}{\mathcal{K}}
\newcommand{\cY}{\mathcal{Y}}
\newcommand{\cX}{\mathcal{X}}
\newcommand{\cV}{\mathcal{V}}
\def\bgamma{\boldsymbol\gamma}

\newcommand{\specialcell}[2][c]{%
  \begin{tabular}[#1]{@{}c@{}}#2\end{tabular}}

\renewcommand{\b}{\ensuremath{\mathbf}}

\def\mc{\mathcal}
\def\mb{\mathbf}

\newcommand{\T}{^{\raisemath{-1pt}{\mathsf{T}}}}

\makeatletter
\DeclareRobustCommand\onedot{\futurelet\@let@token\@onedot}
\def\@onedot{\ifx\@let@token.\else.\null\fi\xspace}
\def\eg{e.g\onedot} \def\Eg{E.g\onedot}
\def\ie{i.e\onedot} \def\Ie{I.e\onedot}
\def\cf{cf\onedot} \def\Cf{Cf\onedot}
\def\etc{etc\onedot} \def\vs{vs\onedot}
\def\wrt{wrt\onedot}
\def\dof{d.o.f\onedot}

\newcommand{\etal}{\textit{et al.}}
\newcommand{\figref}[1]{Fig.~\ref{#1}}
\newcommand{\secref}[1]{Section~\ref{#1}}
\newcommand{\appref}[1]{Appendix~\ref*{#1}}
\newcommand{\algref}[1]{Algorithm~\ref{#1}}
\newcommand{\equref}[1]{Eq.~\ref{#1}}
\newcommand{\tabref}[1]{Table~\ref{#1}}

\newcommand\blfootnote[1]{%
\begingroup
\renewcommand\thefootnote{}\footnote{#1}%
\addtocounter{footnote}{-1}%
\endgroup
}

\renewcommand\UrlFont{\color{black}\rmfamily}

\newcommand*\rot{\rotatebox{90}}
\newcommand{\boldparagraph}[1]{\vspace{0.2cm}\noindent{\bf #1:} }


\newif\ifcomment
\commenttrue


\newcommand{\tocite}[1]{\textcolor{red}{[TOCITE]}}
\newcommand{\todo}[1]{\textcolor{red}{[TODO: #1]}}

\section{Introduction}
3D content editing is a fast-growing research area 
with significant potential to shape the future of digital media.
Recently, NeRF~\cite{mildenhall2020nerf} and its extended methods~\cite{barron2021mip, barron2022mip3,li2020neural, park2021nerfies, peng2021neural} have made it easy to reconstruct static and even dynamic 3D scenes.
Therefore, research on 3D scene editing based on NeRFs has also emerged. 
Many methods~\cite{liu2021editing,yang2022neumesh,huang2022stylizednerf} were proposed for editing static 3D scenes, which provide friendly editing tools for non-professional users. 
However, there is a noticeable lack of research related to editing the appearance of dynamic 3D scenes, despite a great demand for users to add their desired content  to volumetric videos.

This paper specifically focuses on fine-grained local appearance editing for 3D dynamic scenes.
Given a dynamic NeRF (probably in various representations) and its original training videos,
we aim to edit the appearance of the dynamic NeRF by modifying a single 2D frame in the training videos,
as illustrated in~\figref{teaser}.
This problem is challenging for three reasons.
First, it is non-trivial to propagate the edited 2D pixels to the implicit neural scene representation, especially for dynamic scenes.
Moreover, how to propagate the single-frame edited content to other frames in a temporally consistent manner is not well explored.
Finally, designing a generally applicable editing tool that can be plugged into different variants of dynamic NeRF representations ~\cite{peng2021neural,li2020neural,park2021nerfies,Gao2021DynNeRF, li2022neural} presents a significant challenge.
A naive baseline is to directly fine-tune a dynamic NeRF on the edited image.
However, training the dynamic NeRF directly using a single image tends to cause the network to overfit to a single view, leading to the degradation of the rendering quality on other views and the failure of propagating the edited result to other frames, despite the tedious computation.

To tackle these challenges, we propose a novel framework called Dyn-E, 
which enables users to locally edit the appearance of dynamic NeRFs by modifying a single image.
Our approach first lifts the edited local region to 3D space to form a local surface and then uses an invertible network to represent the motion of the local surface, allowing us to propagate the edited results across video frames.
Specifically, our local surface is a textured mesh lifted from the edited region of the single image through the rendered depth map of the given dynamic NeRF. We convert the textured mesh into a local density and color field, which can be rendered together with the given dynamic NeRF.
To propagate the edits to other frames, an invertible motion representation is learned from the input videos, enabling efficient warping of the local surface to different time steps.

Our proposed local surface representation is an independent layer that does not make assumptions about the underlying scene structure, thus it is versatile and can be inserted into most existing dynamic NeRF representations.
Thanks to the local surface representation, we are able to maintain the rendering performance of the dynamic NeRFs outside the edited region.
The surface-based representation also allows us to leverage smoothness and  photometric    
constraints on the surface deformation to regularize the learning of the invertible motion network, which makes the spatial positions of the local surface in all frames more accurate.
We conduct extensive experiments on three commonly used dynamic scene datasets to verify the effectiveness of our algorithm.
Additionally, we demonstrate that our method is suitable for most dynamic NeRFs by editing HyperNeRF~\cite{park2021hypernerf}, DynamicNeRF~\cite{Gao2021DynNeRF}, and Neural Body~\cite{peng2021neural}.

Our main contributions are summarized as follows:
(1) We propose a novel approach to the task of image-based local appearance editing for dynamic NeRFs.
(2) We design a trackable local surface representation that facilitates spatio-temporally consistent dynamic scene appearance editing. 
(3) We extensively verify the effectiveness and versatility of our approach with various dynamic NeRF representations and multiple complex datasets.

\section{Related Works}
\paragraph{Dynamic NeRFs.}
Recently, Neural Radiance Fields (NeRFs)
~\cite{mildenhall2020nerf} demonstrate remarkable performance in the domain of novel view synthesis.
Given a set of multi-view images of static scenes, NeRF has the ability to reconstruct scenes and generate free-viewpoint videos.
Some studies~\cite{pumarola2020d, tretschk2021nonrigid,park2021nerfies,park2021hypernerf, peng2023representing, lin2022efficient} extend NeRFs to dynamic scenes, showcasing promising results.
D-NeRF~\cite{pumarola2020d} and Nerfies~\cite{park2021nerfies} employ canonical NeRFs and a series of deformation fields to capture dynamic scenes.
NSFF~\cite{li2020neural} uses MLPs to model the scene flow fields, ensuring temporal consistency.
DyNeRF~\cite{li2022neural} directly utilizes time-conditioned neural radiance fields for representing dynamic scenes.
Several works~\cite{peng2021animatable, peng2021neural, zheng2022structured, liu2021neural} utilize prior knowledge of the human body to handle dynamic human bodies with a wide range of motion.
Despite achieving high-quality rendering results, these works do not allow users to freely edit their appearance.

\paragraph{Neural scene editing.}
Scene editing based on neural fields is gaining increasing attention.
Some works~\cite{xu2022deforming, yuan2022nerf, liu2021editing, bao2023sine} are able to modify the geometry of scenes, producing impressive results, while others focus on editing the appearance of scenes, similar to our work.
A series of works~\cite{zhang2022arf, huang2022stylizednerf, nguyen2022snerf} utilize pretrained 2D neural networks to modify the style of neural radiance fields.
Some works~\cite{zhang2021nerfactor, zhang2022modeling, zhang2021physg} enable physics-based relighting and material editing on NeRFs by recovering the material properties of objects and the environmental lighting conditions.
Another line of work~\cite{yang2022neumesh, xiang2021neutex, das2022learning} focuses on fine-grained appearance editing of NeRFs.
NeuMesh~\cite{yang2022neumesh} empowers users to modify the appearance of objects by editing a 2D image captured from a specific viewpoint and achieves consistent rendering of the edited appearance from multiple viewpoints.
However, these methods only enable editing the appearance of static scenes and do not explore editing dynamic scenes.
Recently, some works~\cite{ho2023custom, jafarian2023normal, chen2022uv} attempt to edit dynamic human bodies, 
but they need to utilize prior knowledge of the human body which is not required in our work.

\paragraph{Video editing.}
Achieving consistent video editing~\cite{yu2023videodoodles, molad2023dreamix, qi2023fatezero, liu2023video} has always been a long-standing challenge in the field of video editing.
To achieve temporal consistency, some studies~\cite{ruder2016artistic, xu2022temporally} utilize optical flow as a constraint.
Other studies~\cite{jamrivska2019stylizing, texler2020interactive} utilize keyframe propagation methods to propagate modifications made in keyframes to other frames.
Additionally, there are approaches~\cite{kasten2021layered, bar2022text2live, ye2022deformable} that represent videos as 2D atlases, enabling appearance editing of the videos by modifying the atlases.
However, these methods lack 3D awareness, making it non-trivial to extend them for performing novel view synthesis.
Some traditional methods~\cite{deng2022survey, xu2014nonrigid, habermann2019nrst} attempt to edit the appearance of dynamic videos by reconstructing the explicit 3D representation of dynamic objects.
However, they only show the editing results
for simple scenes or in the presence of RGBD training data.

\section{Method}
\begin{figure*}[htbp]

\begin{center}
\includegraphics[width=\textwidth]{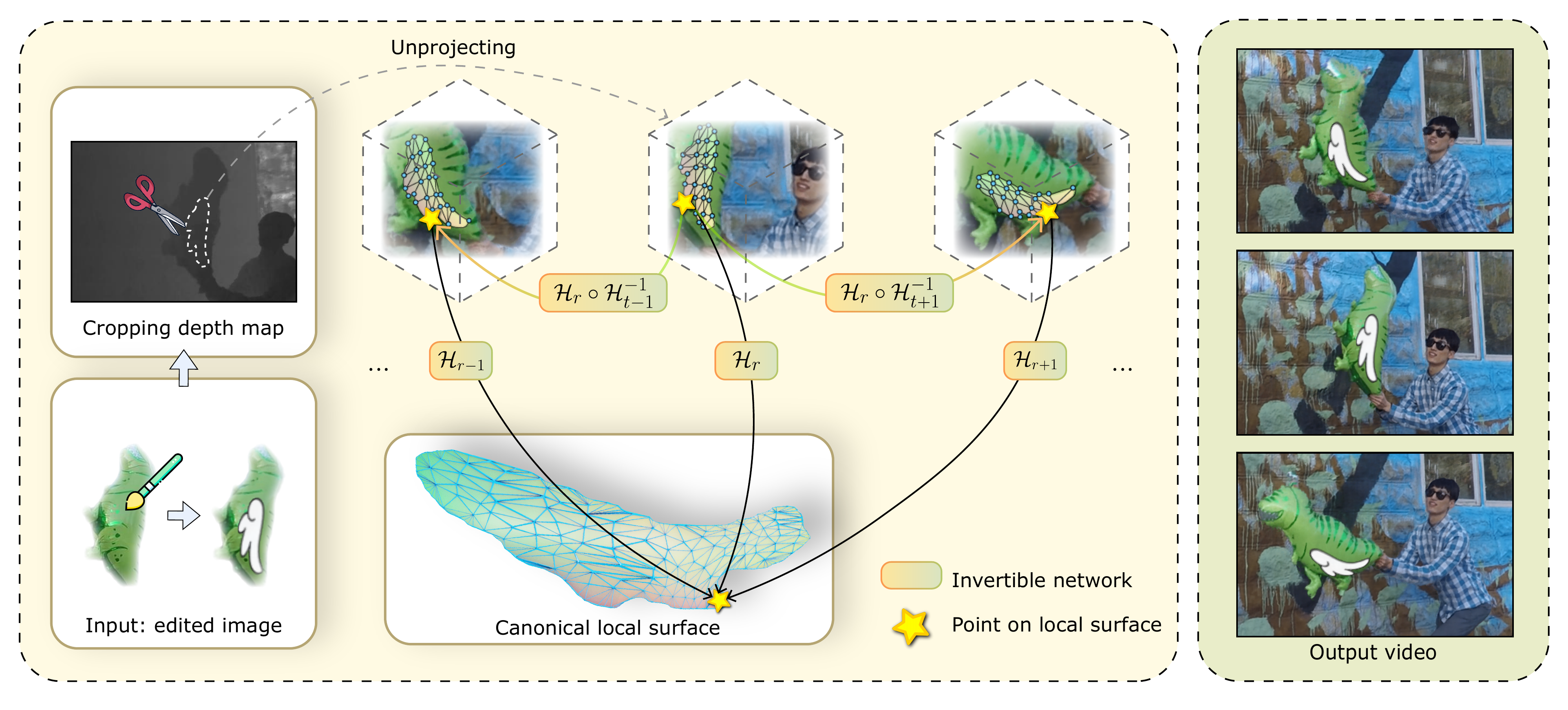}

\end{center}
\vspace{-1em}

\caption{
    \textbf{Illustration of our pipeline.} 
    Given a single edited image and a dynamic NeRF, we first lift the edited region to the 3D space through rendered depth maps to form a textured mesh. Then, we train an invertible network to propagate the textured mesh to other frames. Finally, we combine the textured mesh with the original dynamic NeRF and render them to obtain the final results. 
}
\label{pipeline}
\vspace{-1em}

\end{figure*}

Given a dynamic NeRF and its training video data, our task is to edit the local appearance of the dynamic NeRF by modifying a single 2D image in the training video.
To this end, we propose a framework that can edit the appearance of dynamic NeRF 
by lifting the 2D edited content to a sequence of temporally consistent 3D edited content, as illustrated in~\figref{pipeline}.
We first briefly introduce dynamic NeRFs for modeling dynamic scenes and our problem setting in~\secref{dynerf}.
Then, we describe a local surface representation in~\secref{local_surface}.
Meanwhile, we propose a motion representation that allows for the tracking of the local surface in~\secref{deformation}.
Next, \secref{reg} describes our training losses for the motion representation and the constraints that are applied to the local surface.
Finally, \secref{editing} illustrates how to render edited results after training.

\subsection{Preliminary}
\label{dynerf}
\paragraph{Dynamic NeRFs.}
In this paper, we use the term ``dynamic NeRF'' to refer to the neural radiance fields that can represent dynamic scenes.
Existing dynamic NeRFs~\cite{park2021nerfies, peng2021neural, park2021hypernerf,Gao2021DynNeRF,li2022neural,li2020neural} represent dynamic scenes as a time-varying continuous neural radiance field,
i.e., a mapping function $f_\theta$ from a spatial position $\mathbf{x}$, time $t$, and viewing direction $\mathbf{d}$ to color and density.
It can be represented as a general equation:
\begin{equation}
    f_\theta:\left(\mathbf{x} \in \mathbb{R}^3, t \in \mathbb{R}, \mathbf{d} \in \mathbb{S}^2\right) \mapsto\left(\sigma \in \mathbb{R}^{+}, \mathbf{c} \in \mathbb{R}^3\right).
\end{equation}
Different dynamic NeRFs have different ways of modelling time-varying scene content.
Some works~\cite{Gao2021DynNeRF,li2020neural} adopt MLPs taking an additional ``time coordinate'' as input to represent the dynamic components of the scene.
Nerfies~\cite{park2021nerfies} uses a series of time-varying deformation fields to represent the change of scene over time.
Neural Body~\cite{peng2021neural} represents the dynamic content of the scene by a time-varying human parametric model.
Our method is designed to be compatible with most dynamic NeRFs.

\paragraph{Problem setting.}
Our goal is to edit the appearance of the given dynamic NeRF $f_\theta$ by modifying a single 2D image in the training video  $\{\mathbf{I}_i\}^N_{i=1}$, 
and then render a free-viewpoint video that is temporally consistent and high-quality.
For convenience, we refer to the image that the user wants to edit as the reference image. 
Our method can be trained on both monocular and multi-view data. $N_v$ denotes the number of cameras.

We assume that the region to be edited can be observed without occlusion in most cases.
Additionally, we assume that the user only edits part of the image, rather than the entire image.

\subsection{Local Surface}
\label{local_surface}
It is difficult to edit the appearance of dynamic NeRFs by finetuning them on a single image.
This is because optimizing dynamic NeRFs on a single image tends to cause overfitting problems.
Most dynamic NeRFs use an MLP to represent the entire space, thus finetuning their local regions will also affect other regions.
For local appearance editing tasks, we observe that most 3D regions do not need to be edited. 
Hence,  we only need to design a local layer to be inserted into the dynamic NeRF. By doing so, we can solely modify the parameters of this local layer, ensuring that other regions remain unaffected.
Motivated by this, we propose a plug-and-play local surface representation to edit the appearance of dynamic NeRFs.
Our local surface representation adopts a mesh-based surface representation that can be easily constructed.
Moreover, we design our surface representation to be compatible with the volume rendering equation to handle occlusion.
To generate the local surface, we first render the depth map for the reference image using the given dynamic NeRF.
Then, we unproject the user-edited region of the reference image back to 3D space to form the mesh, similar to Shih~\etal~\cite{Shih3DP20}.
Specifically, we unproject user edited pixels $\{\mathbf{p}_i\}^K_{i=1}$ back to 3D space in world coordinates to form the vertices $\{\mathbf{v}_r^i\}^K_{i=1}$ of the mesh, as shown in~\equref{unproject}:
\begin{equation}
\{\mathbf{v}_r^i\}^K_{i=1} = \mathbf{M}_{c2w} \Pi^{-1}(\{\mathbf{p}_i\}^K_{i=1}),
\label{unproject}
\end{equation}
where $\Pi^{-1}$ represents the inverse perspective projection operation. $\mathbf{M}_{c2w}$ is the camera-to-world transformation matrix, 
which is used to transform the mesh vertices from camera coordinates to world coordinates.
$K$ is the number of mesh vertices. 
The vertices of neighboring pixels are connected to form the faces of the mesh.

We define the color of each vertex of the mesh as the color of the corresponding pixel in the reference image.
When rendering the mesh, we manually define the color for each point along the ray.
For any point on the rays intersecting the mesh, its color $\mathbf{c}^s$ is calculated by interpolating the vertex colors of the intersecting face of the 3D mesh using barycentric coordinates as weights.
For any point on the rays that does not intersect the mesh, its color $\mathbf{c}^s$ is set to zero.

Although the local surface representation is easy to construct,
it is difficult to handle occlusion relationships between the local surface and the original dynamic NeRF, as shown in~\figref{occ}.
To address this problem, we  design our surface representation to be seamlessly integrated with the given dynamic NeRF for volume rendering.
We convert the mesh into a distance field according to the nearest distance from a point to the mesh.
Then, the distance field is converted into a density field.
Inspired by VolSDF~\cite{yariv2021volume}, we convert the distance field into the density field using~\equref{volsdf}:
\begin{equation}
    \sigma^d(\mathbf{x})=\alpha \Psi_\beta\left(-d(\mathbf{x})\right),
    \label{volsdf}
\end{equation}
where $\alpha=\beta^{-1}$, $\beta$ is a 
 manually defined parameter, 
$d(\mathbf{x})$ is the distance from the mesh to the point $\mathbf{x}$,
and $\Psi_\beta$ represents the cumulative distribution function of the Laplace distribution with zero mean and a scale of $\beta$.
To ensure that the local surface replace the original dynamic NeRF in the edited region 
and does not affect other regions of the dynamic NeRF,
we define a mask field to indicate which regions belong to the local surface and which regions belong to the dynamic NeRF.
For points within a distance of $\gamma$ from the surface, the mask value is set to 1. For points beyond a distance of $\gamma$ from the surface, the mask value is set to 0.
If a ray $\mathbf{r}(t)$ does not intersect with the mesh, we set the mask value to 0 for all points on the ray.
The mask field is defined as:
\begin{equation}
M(\mathbf{x}) = \begin{cases}
    1, & \text{if $ d(\mathbf{x}) < \gamma$ and $\mathbf{r}(t) \in \mathcal{R}_h$} \\
    0, & \text{otherwise}
\end{cases},
\label{mask}
\end{equation}
where $\mathcal{R}_h$ is the set of all rays that intersect with the mesh.
Our ultimate volume rendering equation is:
\begin{equation}
    \mathbf{C}^{full}(\mathbf{r})=\sum_{i=1}^N T_i^{full}\left(\alpha\left(\sigma_i^d \delta_i\right)(1-M_i) \mathbf{c}_i^d+\alpha\left(\sigma_i^s \delta_i\right) M_i \mathbf{c}_i^s\right),
    \label{volume}
\end{equation}
where $\mathbf{C}^{full}$ is the rendered color of the ray $\mathbf{r}$ and
$\alpha(x)=1-\exp \left(-x\right)$.
$\sigma^d$ and $\sigma^s$ are the densities of the dynamic NeRF and the local surface,
$\delta$ is the distance between adjacent sample points.
$M_i$ is the mask value at the $i$-th sample point. $\mathbf{c}^d$ and $\mathbf{c}^s$ are the colors of the dynamic NeRF and the local surface.
$T_i^{full} = \exp(-\sum_{j=1}^{i-1} (\sigma_j^d \delta_j (1-M_j) + \sigma_j^s \delta_j M_j))$
is the accumulated transmittance of the ray $\mathbf{r}$ up to the $i$-th sample point.

\subsection{Motion Representations}
\label{deformation}
Through the above method, we only generate the local surface of a certain frame.
The positions of the local surface of other frames are still unknown.
Therefore, we propose our motion representation to propagate the local surface to other frames.
Inspired by CaDeX~\cite{Lei2022CaDeX}, we use invertible networks to model our scene motion.
Invertible networks are strictly bijective maps, which fits the natural properties of non-rigid motion better than the traditional method~\cite{tewari2022disentangled3d} of using two MLPs to represent the deformation and the inverse deformation.
By leveraging the invertible network, we can know the point-to-point correspondence between arbitrary two frames, as shown in~\equref{invwarp}:
\begin{equation}
    \mathbf{x}_{t \rightarrow t'}=\mathcal{H}_{t'}^{-1} \circ \mathcal{H}_t\left(\mathbf{x}_t\right),
\label{invwarp}
\end{equation}
where $\mathcal{H}_t$ represents the operation of mapping the point $\mathbf{x}_t$ from the the observation space at frame $t$ to the canonical space by the invertible network,
and $\mathcal{H}_{t'}^{-1}$ is the inverse operation of mapping the point $\mathbf{x}_{t'}$ from the canonical space to the observation space at frame $t'$.

Theoretically, we can train the invertible network to learn the correspondence between adjacent frames in the same way as NSFF~\cite{li2020neural}.
Because the invertible network is a strictly bijective map, after learning the correspondence between each pair of adjacent frames,
the correspondence between any two frames is naturally known.
However, this training strategy needs to 
densely sample the points in 3D space to transform the scene flows into optical flows on the 2D image plane for 2D supervision.
Due to the large GPU memory consumption of the invertible network, 
we cannot use the same strategy as above to train the invertible network.
To resolve this problem, we employ an MLP $f_\phi$ to model the scene flow field and then use $f_\phi$ to distill the motion information to the invertible network.
The MLP $f_\phi$ is defined as:
\begin{equation}
\left(\mathbf{f}_{t \rightarrow t+1}, \mathbf{f}_{t \rightarrow t-1}\right)=f_\phi(\mathbf{x}, \gamma(t)), 
\label{scene_flow}
\end{equation}
where $\mathbf{f}_{t \rightarrow t+1}$ and $\mathbf{f}_{t \rightarrow t-1}$ are the scene flow fields from frame $t$ to frame $t+1$ and frame $t$ to frame $t-1$,
 and $\gamma(t)$ is the positional encoding of $t$.
We adopt the loss function described in~\equref{invertible_networks} to train the invertible network and only sample training points near the local surface, which can greatly reduce memory consumption.
\begin{equation}
\mathcal{L}_{distill}=\sum_{\mathbf{x}\in \mathcal{X}_{surf}}(||\mathbf{f}_{t \rightarrow t+1}^{\mathrm{inv}}-\mathbf{f}_{t \rightarrow t+1}||+||\mathbf{f}_{t \rightarrow t-1}^{\mathrm{inv}}-\mathbf{f}_{t \rightarrow t-1}||),
\label{invertible_networks}
\end{equation}
where $\mathbf{f}_{t \rightarrow t+1}^{\mathrm{inv}} = \mathbf{x}_{t \rightarrow t+1} - \mathbf{x}_t$ and $\mathbf{f}_{t \rightarrow t-1}^{\mathrm{inv}} = \mathbf{x}_{t \rightarrow t-1} - \mathbf{x}_t$ 
are the scene flows predicted by the invertible networks. 
$\mathcal{X}_{surf}$ is the set of points near the local surface.
\subsection{Training}
\label{reg}
We utilize the motion matching loss to supervise the scene flow field $f_\phi$:
\begin{equation}
\label{motion_matching_loss}
\mathcal{L}_{motion}=\sum_{\mathbf{r}\in \mathcal{R}} \sum_{j \in\{i \pm 1\}} || \hat{\mathbf{p}}_{i \rightarrow j}\left(\mathbf{r}_i\right)-\mathbf{p}_{i \rightarrow j}\left(\mathbf{r}_i\right) ||,
\end{equation}
where $\mathcal{R}$ is the set of rays, $\hat{\mathbf{p}}_{i \rightarrow j}$ 
is the ground truth optical flow from frame $i$ to frame $j$ predicted by RAFT~\cite{teed2020raft},
and $\mathbf{p}_{i \rightarrow j}$ is the optical flow induced by the scene flow $\mathbf{f}_{i \rightarrow j}$.

The deformation sequence of our local surface can be recovered easily through the invertible networks,
which allows us to easily add regularization terms on the local surface to eliminate accumulated errors caused by the scene flow.
We introduce the Laplacian smooth term and the feature-based photometric term as follows.

The Laplacian smooth term is a common regularization term used to make the mesh more smooth, as shown in~\equref{laplacian_smooth}:
\begin{equation}
\mathcal{L}_{lap}=\sum_{\mathbf{v}^i \in \mathcal{V}}  \left\|\mathbf{v}^i_t-\mathbf{\bar{v}}^i_t\right\|^2 \text{, where } 
\mathbf{\bar{v}}^i_t = \sum_{j \in \mathcal{N}(i)} \mathbf{v}^j_t,
\label{laplacian_smooth}
\end{equation}
where $\mathcal{N}(i)$ is the set of neighboring vertices of the $i$-th vertex, 
and $\mathbf{v}^i_t$ is the $i$-th vertex of the local surface at frame $t$. $\mathcal{V}$ is the set of all the vertices of the local surface.
To align the surface with the image in each frame, 
the feature-based photometric term is utilized to measure the alignment between the surface and the image,
as shown in~\equref{feature_based_photometric}:
\begin{equation}
\begin{aligned}
&\mathcal{L}_{pho}= \sum_{\mathbf{v}^i \in \mathcal{V}} \sum_{j=0}^{N_v}  s_{\cos }(F_{r}(\Pi(\mathbf{v}_r^i)), F^j_{t}(\Pi(\mathbf{v}_t^i))), \\
&    s_{\cos }(x, y)= max(1 - \frac{\langle x, y\rangle}{\|x\| \cdot\|y\|}, \epsilon),
\label{feature_based_photometric}
\end{aligned}
\end{equation}
where $F_{r}$ and $F_{t}$ are the features extracted from the reference image and the image at frame $t$,
$\Pi$ is the projection function  and $\mathbf{v}_r$ and $\mathbf{v}_t$ are the vertices of the local surface at the reference frame and frame $t$.
$N_v$ is the number of cameras.
$\epsilon$ is a threshold to prevent inaccurate supervision signals caused by occlusion.
We adopt S2DNet~\cite{Germain2020S2DNet} as the feature extractor, for it can extract robust local features.
Note that only $\mathcal{L}_{pho}$ may be supervised by multi-view images, while the other loss functions solely require supervision from single-view videos.
The complete loss functions are shown in the supplementary material.
\subsection{Rendering Edited Results}
\label{editing}
After training, we can warp the local surface to any frame we want by the learned invertible networks. Then, we combine the local surface and the given dynamic NeRF, rendering them together according to~\equref{volume}. 

\section{Experiments}
\subsection{Experimental Settings}
\paragraph{Comparison methods.}
\begin{figure*}
\begin{center}
\setlength\tabcolsep{0.2em}
\newcommand{\mywidth}{0.28 \textwidth}
\begin{tabular}{ccc}
\includegraphics[width=\mywidth]{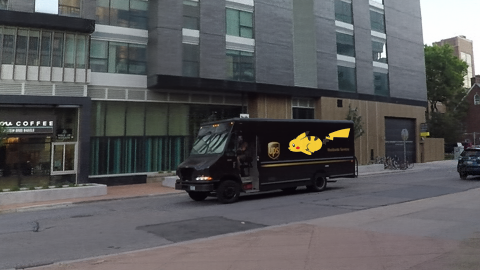}
&\includegraphics[width=\mywidth]{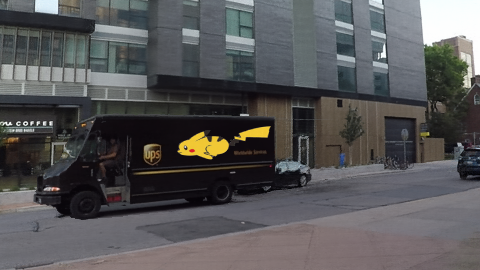}
&\includegraphics[width=\mywidth]{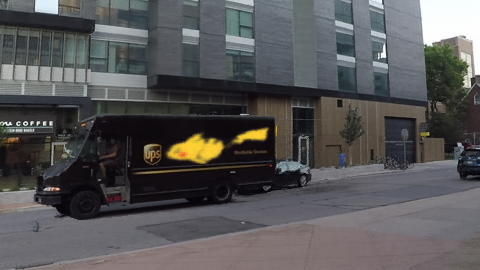}
\\
\textbf{Input: }edited image
&
Ours
& 
SF+DynNeRF
\\[2ex]
\includegraphics[width=\mywidth]{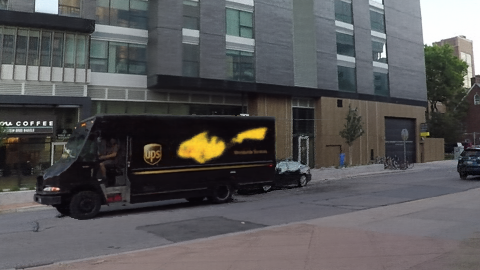}
&\includegraphics[width=\mywidth]{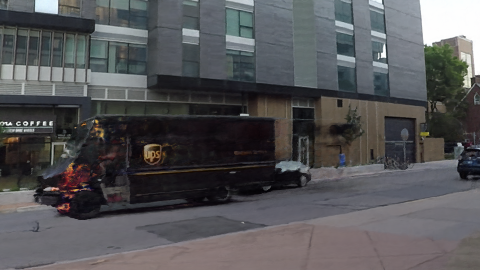}
&\includegraphics[width=\mywidth]{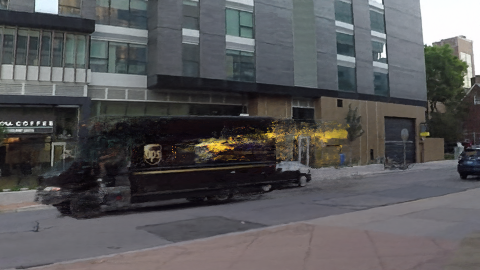}
\\

OF+DynNeRF
&
Nerfies
& 
HyperNeRF
\\[2ex]
\includegraphics[width=\mywidth]{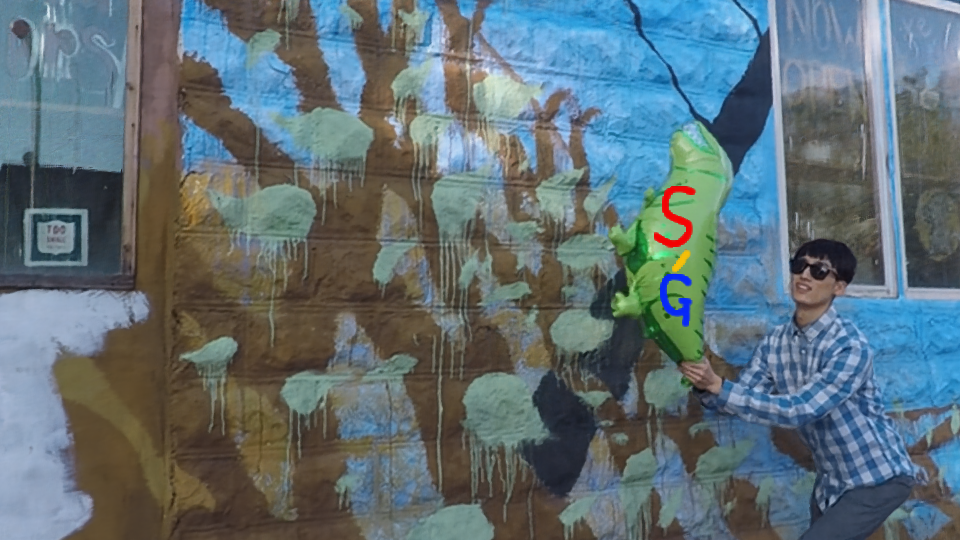}
&\includegraphics[width=\mywidth]{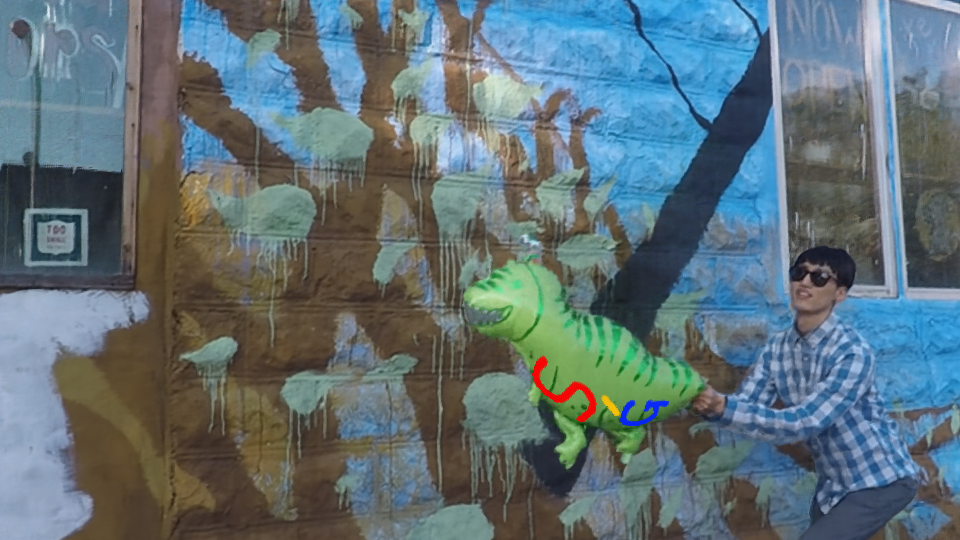}
&\includegraphics[width=\mywidth]{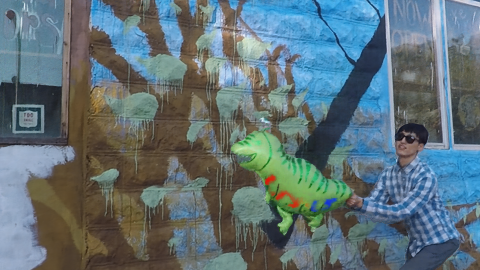}
\\
\textbf{Input: }edited image
&
Ours
& 
SF+DynNeRF
\\[2ex]
\includegraphics[width=\mywidth]{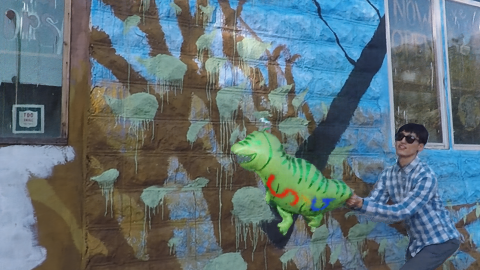}
&\includegraphics[width=\mywidth]{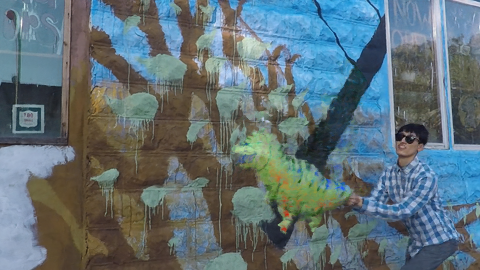}
&\includegraphics[width=\mywidth]{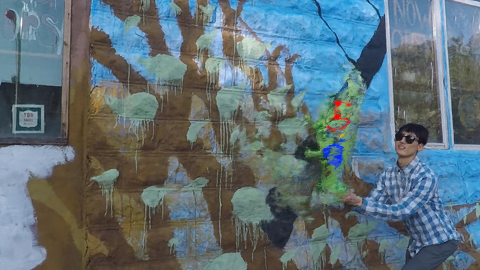}
\\

OF+DynNeRF
&
Nerfies
& 
HyperNeRF
\\[2ex]
\includegraphics[width=\mywidth]{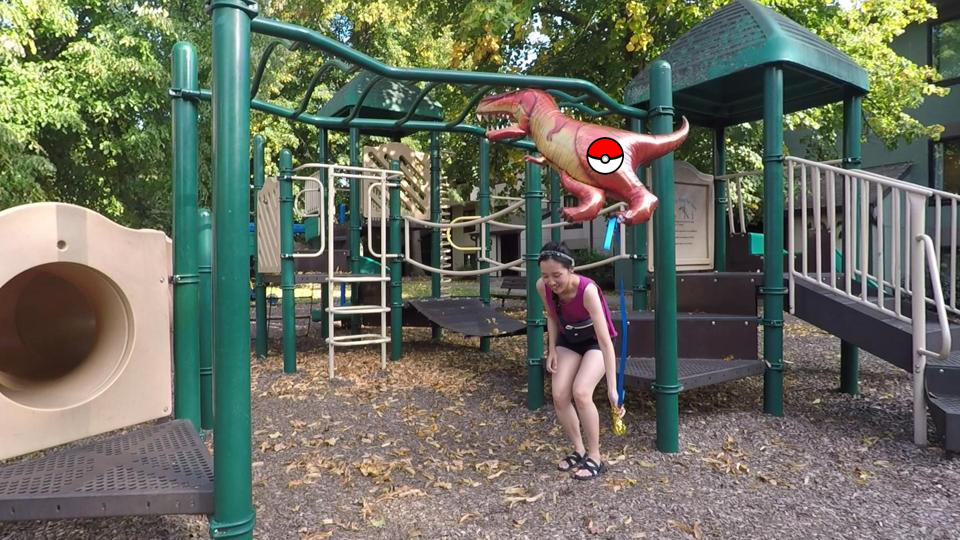}
&\includegraphics[width=\mywidth]{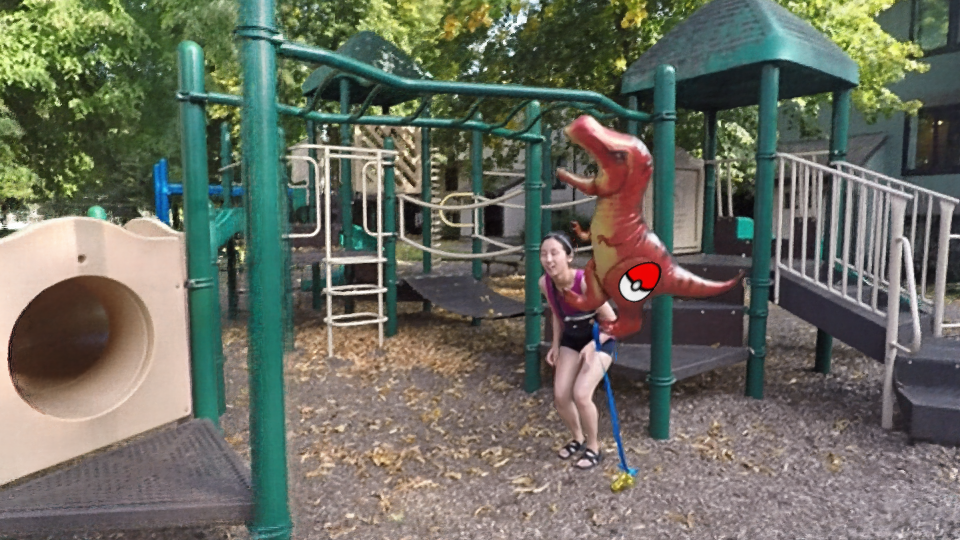}
&\includegraphics[width=\mywidth]{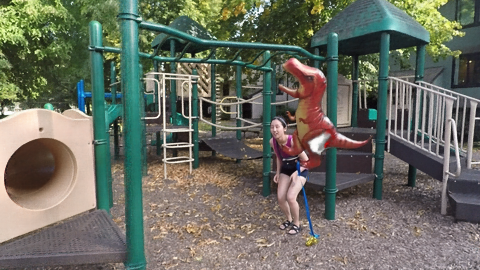}
\\
\textbf{Input: }edited image
&
Ours
& 
SF+DynNeRF
\\[2ex]
\includegraphics[width=\mywidth]{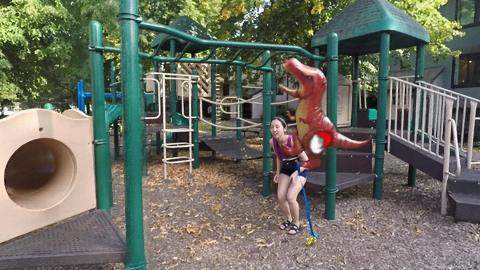}
&\includegraphics[width=\mywidth]{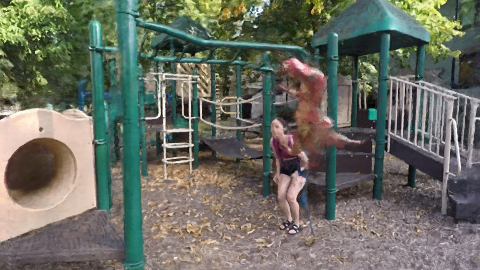}
&\includegraphics[width=\mywidth]{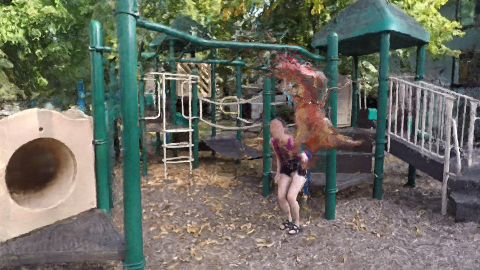}
\\

OF+DynNeRF
&
Nerfies
& 
HyperNeRF
\end{tabular}
\vspace{-1em}
\end{center}

\captionsetup{font={normalsize}} 
\caption{\textbf{Qualitative comparisons.}
We generate more realistic results than the baseline methods.
}
\label{fig:exp}
\end{figure*}

\begin{table}\centering
    \caption{\textbf{User study results.} 
    Our method is preferred by users in terms of temporal consistency and photo-realism.
    }
    \vspace{-0.5em}
    \begin{threeparttable}
    \begin{tabular}{c c c c c}
    
    \toprule
                        & Temporal consistency$\uparrow$    & Photo-realism$\uparrow$  \\
    \midrule
    Ours         & \textbf{3.75}    & \textbf{3.65}    \\
    Nerfies     & 1.50 & 1.55  \\
    HyperNeRF   & 1.20 & 1.30 \\
    SF+DynNeRF       & 2.15             & 2.30       \\
    OF+DynNeRF     & 3.20            & 2.30           \\
    \bottomrule 
    \end{tabular}
    \end{threeparttable}
    \label{usr}
    \vspace{-1em}
\end{table}
We compare our proposed method with the following baseline methods using the Nvidia Dynamic Scenes Dataset~\cite{yoon2020novel}:
(1) Gao~\etal~\cite{Gao2021DynNeRF} is a widely used method for modeling dynamic scenes. 
To support editing, we make the following modifications:
we project 3D scene flows onto 2D space to render optical flows, and use optical flows to warp the edited content to other frames in the training video.
After obtaining the edited training video, we use it to supervise the time-varying dynamic NeRF.
This baseline is abbreviated as ``SF+DynNeRF''.
(2) ``OF+DynNeRF'' is another strong baseline.
We employ optical flows predicted by the RAFT~\cite{teed2020raft} to warp the edited content in the reference image to other frames in the training video 
and use the edited video to supervise the time-varying dynamic NeRF.
We choose Gao~\etal~\cite{Gao2021DynNeRF} as the dynamic NeRF for this baseline.
This baseline is abbreviated as ``OF+DynNeRF''.
(3) Nerfies~\cite{park2021nerfies} is a commonly used algorithm for modeling non-rigidly deforming scenes for monocular videos. 
It models the non-rigidly deforming scene as a canonical space and a set of deformation fields. 
To edit Nerfies, we finetune it using the single edited image.
To prevent overfitting, we freeze the parameters of the deformation fields and only train the MLP representing the canonical space.
(4) HyperNeRF~\cite{park2021hypernerf} 
extends Nerfies and adopts a family of higher-dimensional spaces to handle topological variations.
HyperNeRF is also finetuned on the single edited image for editing. We only train the MLP that represents the canonical space while freezing the parameters of other components.

\paragraph{Datasets.}
We validate our algorithm on three datasets to demonstrate its wide applicability to various types of dynamic scenes.
(1) The Nvidia Dynamic Scenes Dataset~\cite{yoon2020novel} is a dataset containing 9 videos, which is widely used to measure the performance of dynamic view synthesis.
We use the data processed by Gao~\etal~\cite{Gao2021DynNeRF} for comparison experiments.
To reconstruct the dynamic scenes in this dataset for editing, we choose the method proposed by Gao~\etal~\cite{Gao2021DynNeRF}.
Twenty users were asked to edit these videos and then provide ratings for our algorithm and baseline.
We consider two aspects for user evaluations: (1) temporal consistency and (2) photo-realism. The scores are integers ranging from 1 to 5, where 1 represents lower quality and 5 represents the highest quality.
(2) The Nerfies-HyperNeRF-CoNeRF dataset is utilized by Nerfies~\cite{park2021nerfies}, HyperNeRF~\cite{park2021hypernerf}, and CoNeRF~\cite{kania2022conerf} for their respective research.
Since their capture protocols are similar, we treat them as a single dataset. We reconstruct the dynamic scenes using HyperNeRFs on this dataset and apply our method for editing.
(3) The ZJU-MoCap~\cite{shuai2022multinb, peng2021neural} is a dataset widely used to test the novel view synthesis for human bodies.
Following previous work~\cite{shuai2022multinb}, we use eight synchronized cameras as training views.
We adopt the Neural Body~\cite{peng2021neural} to reconstruct the dynamic human body and utilize our method to edit it.
We present our qualitative results and user study results with baselines on the Nvidia Dynamic Scenes Dataset. 
On the Nerfies-HyperNeRF-CoNeRF and ZJU-MoCap datasets, we show qualitative results of our method.
\paragraph{Implementation details.}
We use the Adam optimizer to train our network with a learning rate of 5e-4.
The specific network architecture is in the supplementary material.
Our network needs to be trained for 12 hours on an NVIDIA RTX 3090 for the Nvidia Dynamic Scenes Dataset, and two NVIDIA RTX 3090 GPUs for the Nerfies-HyperNeRF-CoNeRF and ZJU-MoCap datasets.
The detailed parameter selection of $\beta$ and $\gamma$ is in the supplementary material.

\subsection{Experiment Results}
\begin{figure*}
\begin{center}
\setlength\tabcolsep{0.2em}
\newcommand{\mywidth}{0.16 \textwidth}
\begin{tabular}{cccccc}
\includegraphics[width=\mywidth]{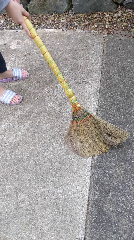}
&\includegraphics[width=\mywidth]{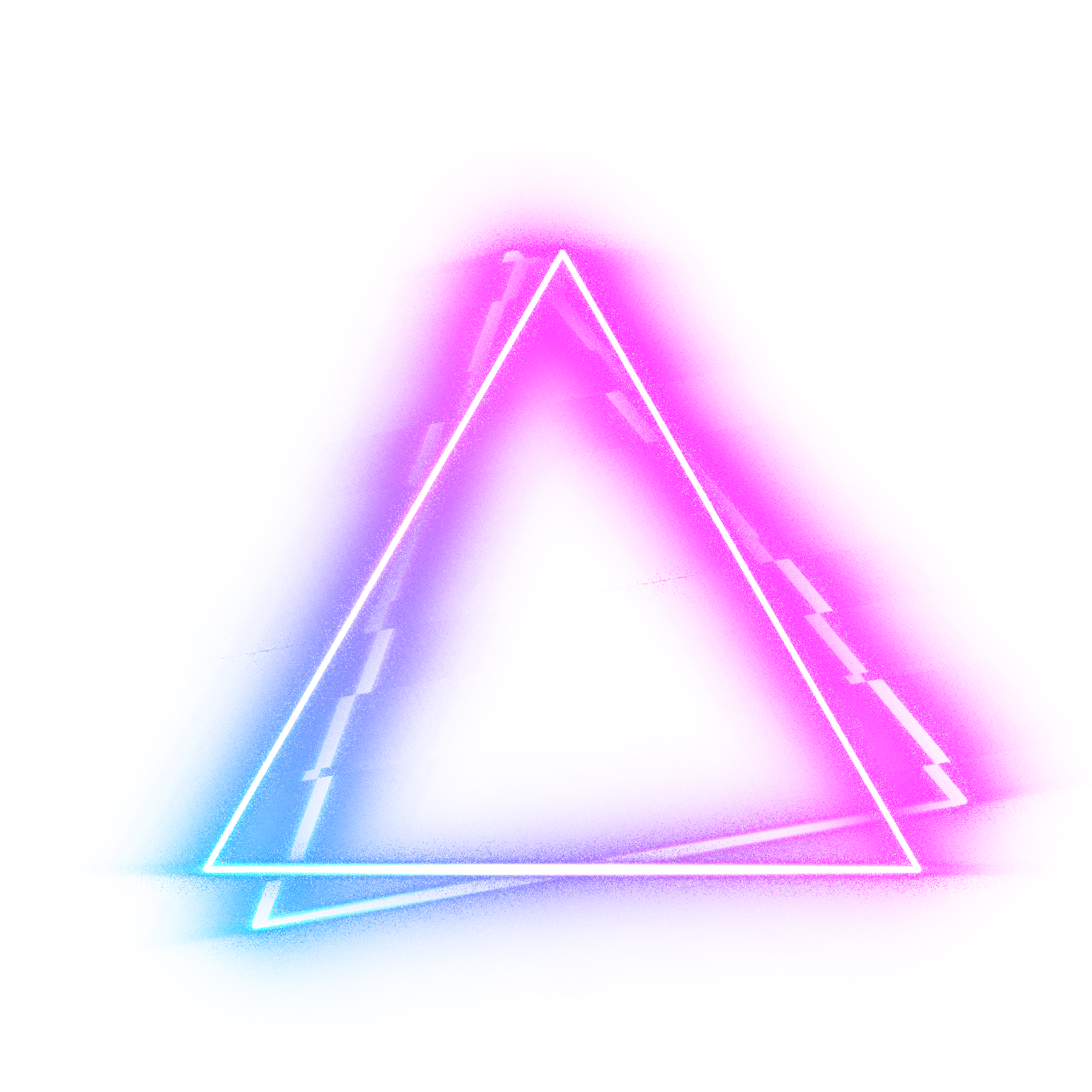}
&\includegraphics[width=\mywidth]{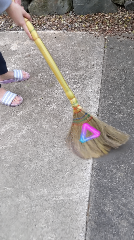}
&
\includegraphics[width=\mywidth]{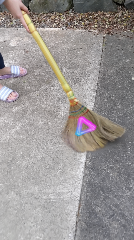}
&
\includegraphics[width=\mywidth]{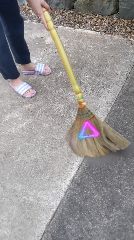}
&\includegraphics[width=\mywidth]{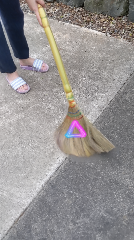}
\\
Original image
&
Added texture
&
Frame 1
& 
Frame 2
&
Frame 3
&
Frame 4
\\[2ex]
\includegraphics[width=\mywidth]{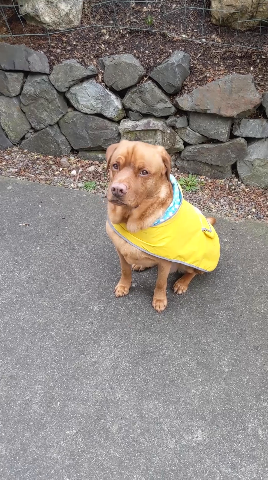}
&\includegraphics[width=\mywidth]{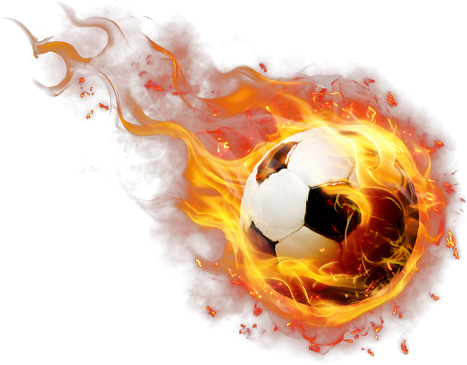}
&\includegraphics[width=\mywidth]{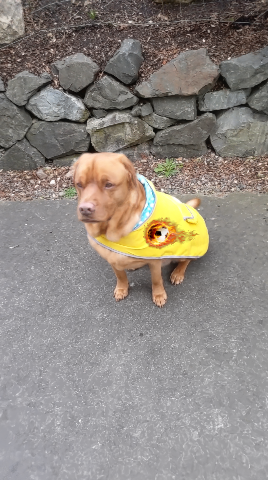}
&
\includegraphics[width=\mywidth]{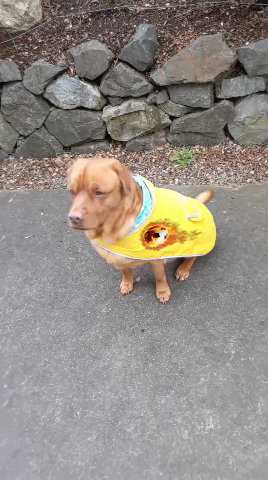}
&\includegraphics[width=\mywidth]{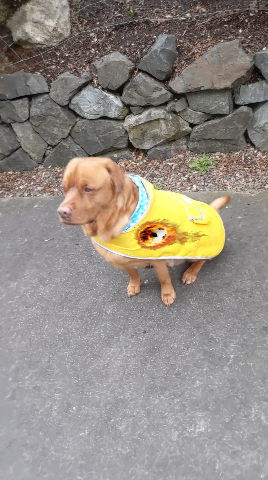}
&\includegraphics[width=\mywidth]{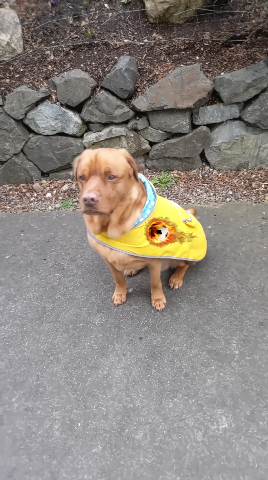}
\\
Original image
&
Added texture
&
Frame 1
& 
Frame 2
&
Frame 3
&
Frame 4
\\[2ex]
\includegraphics[width=\mywidth]{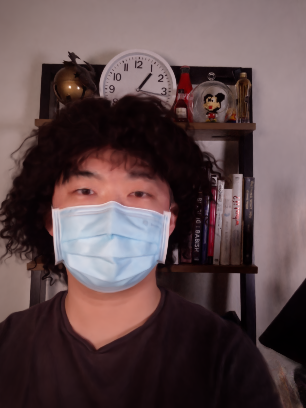}
&\includegraphics[width=\mywidth]{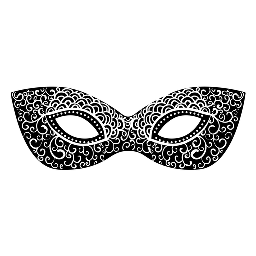}
&\includegraphics[width=\mywidth]{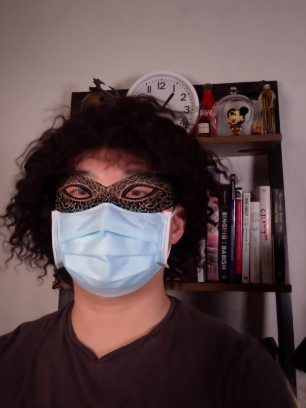}
&
\includegraphics[width=\mywidth]{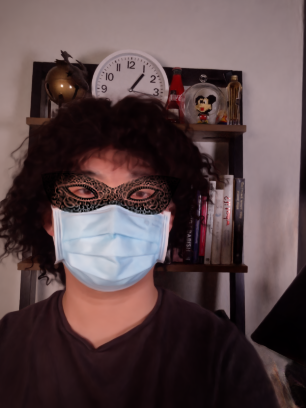}
&\includegraphics[width=\mywidth]{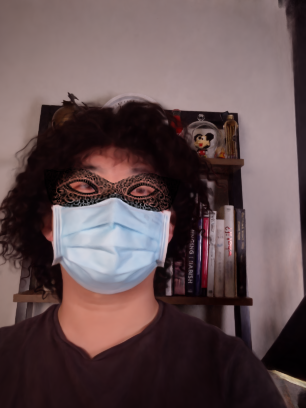}
&\includegraphics[width=\mywidth]{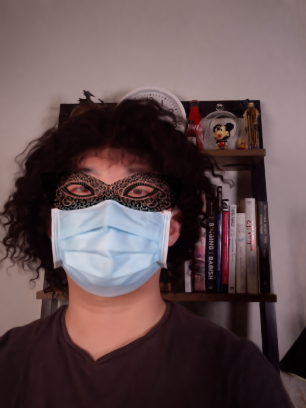}
\\
Original image
&
Added texture
&
Frame 1
& 
Frame 2
&
Frame 3
&
Frame 4
\\
\end{tabular}

\end{center}
\vspace{-0.2em}
\captionsetup{font={normalsize}} 
\caption{\textbf{Qualitative results on the Nerfies-HyperNeRF-CoNeRF dataset.
} More results are shown in the supplementary video.
}
\vspace{-1em}

\label{nerfies}
\end{figure*}

\begin{figure*}
\begin{center}
\setlength\tabcolsep{0.2em}
\newcommand{\mywidth}{0.22 \textwidth}
\begin{tabular}{cccc}
\includegraphics[width=\mywidth]{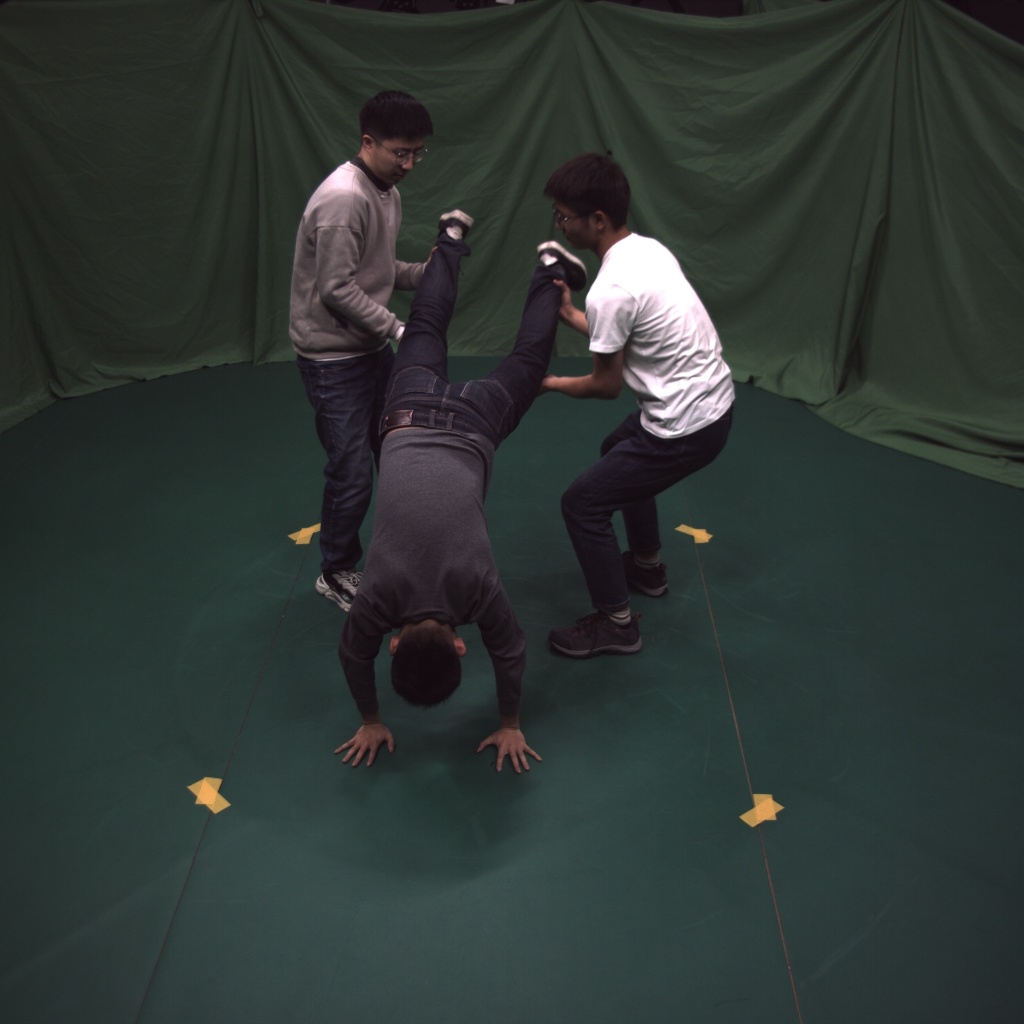}
&\includegraphics[width=0.1 \textwidth]{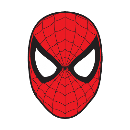}
& \includegraphics[width=\mywidth]{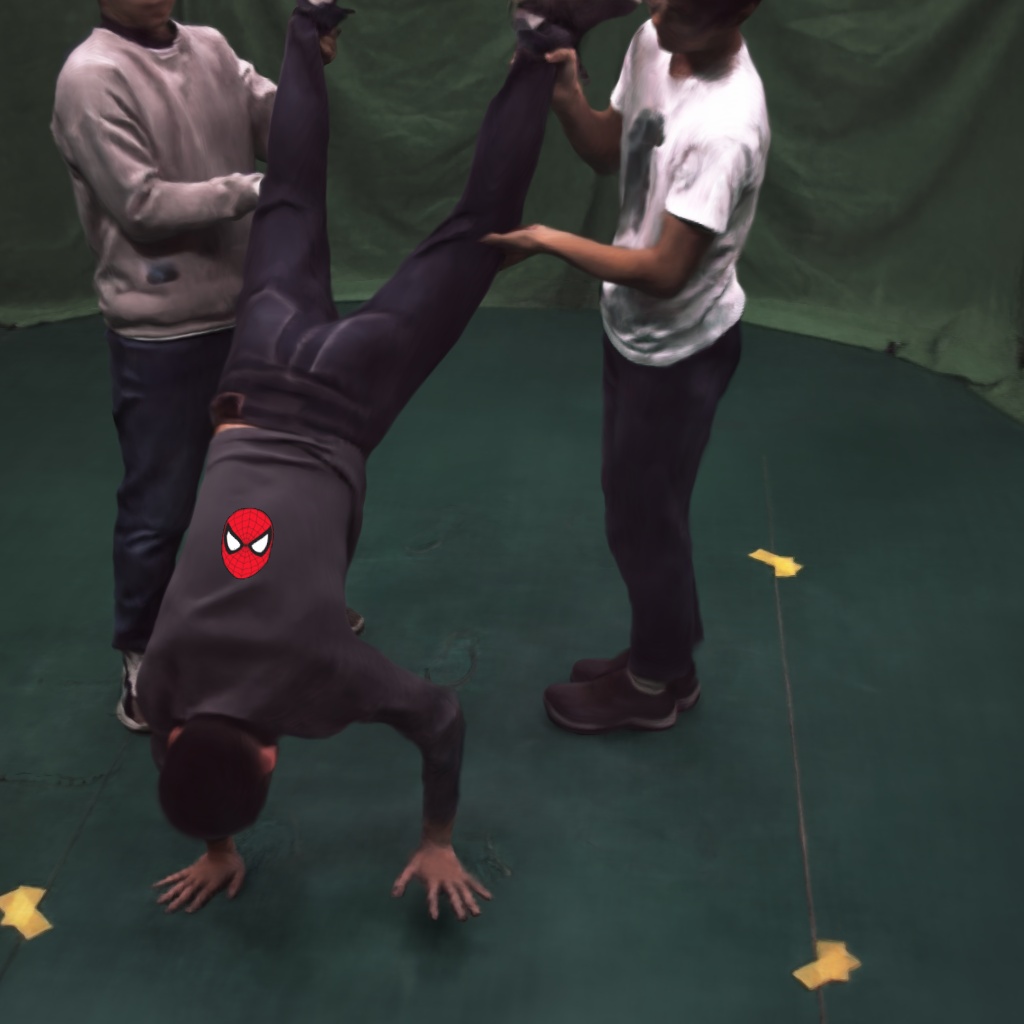}
&\includegraphics[width=\mywidth]{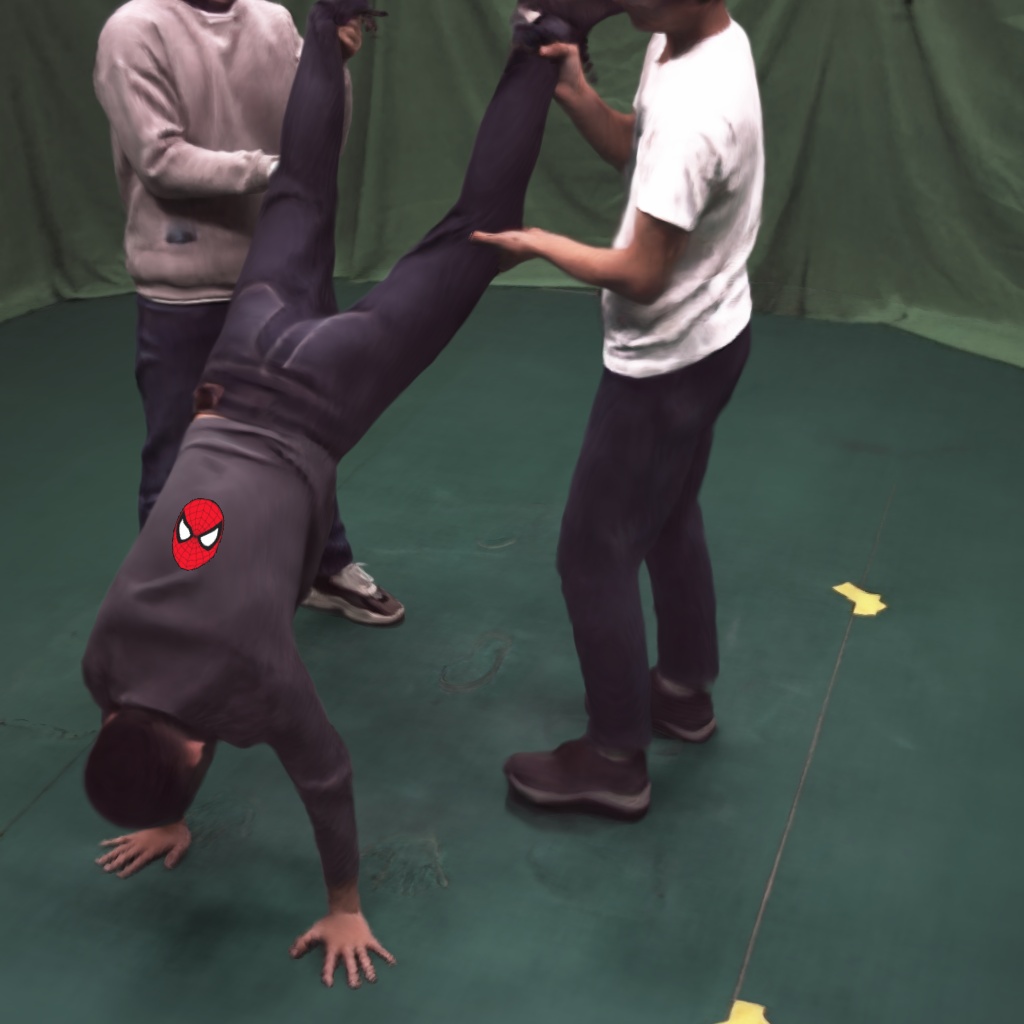}

\\
Original image
&
Added texture
&
Frame 1
& 
Frame 2

\\[2ex]

\includegraphics[width=\mywidth]{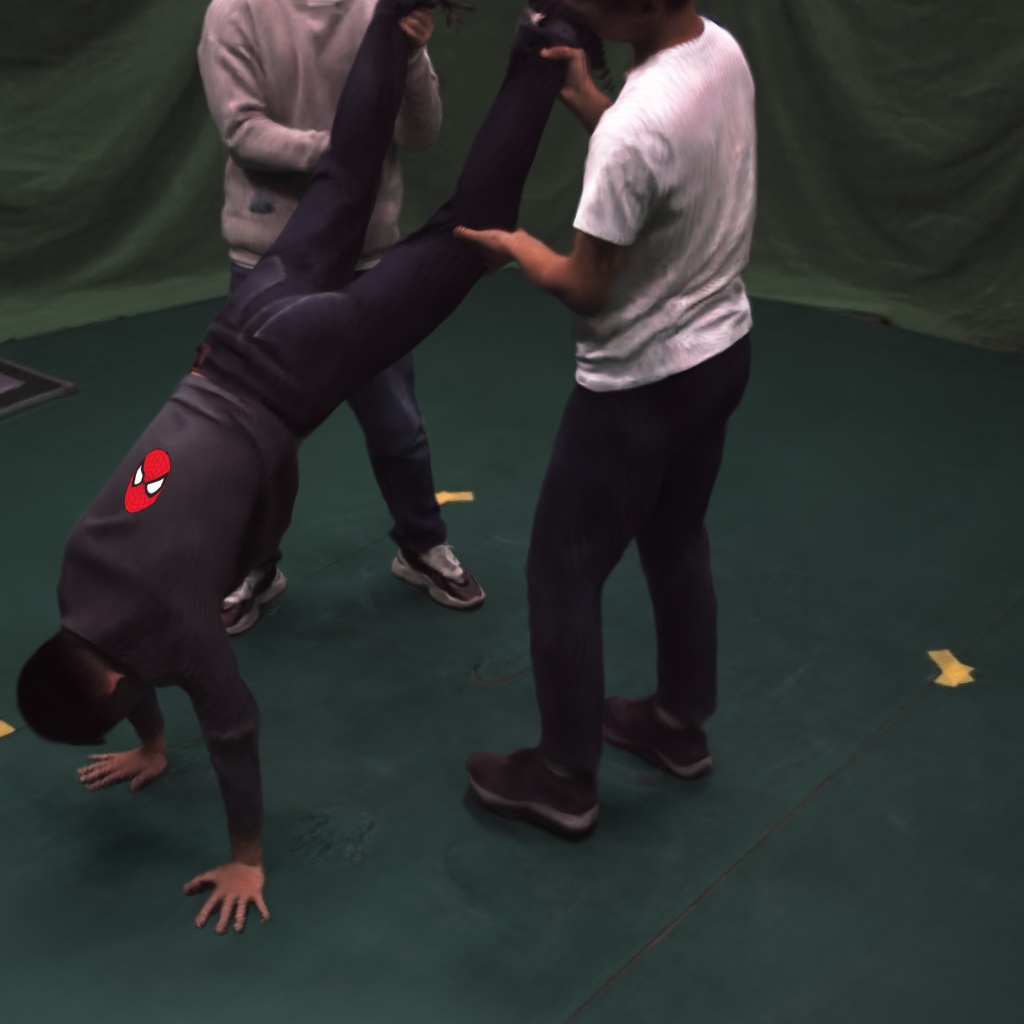}
&
\includegraphics[width=\mywidth]{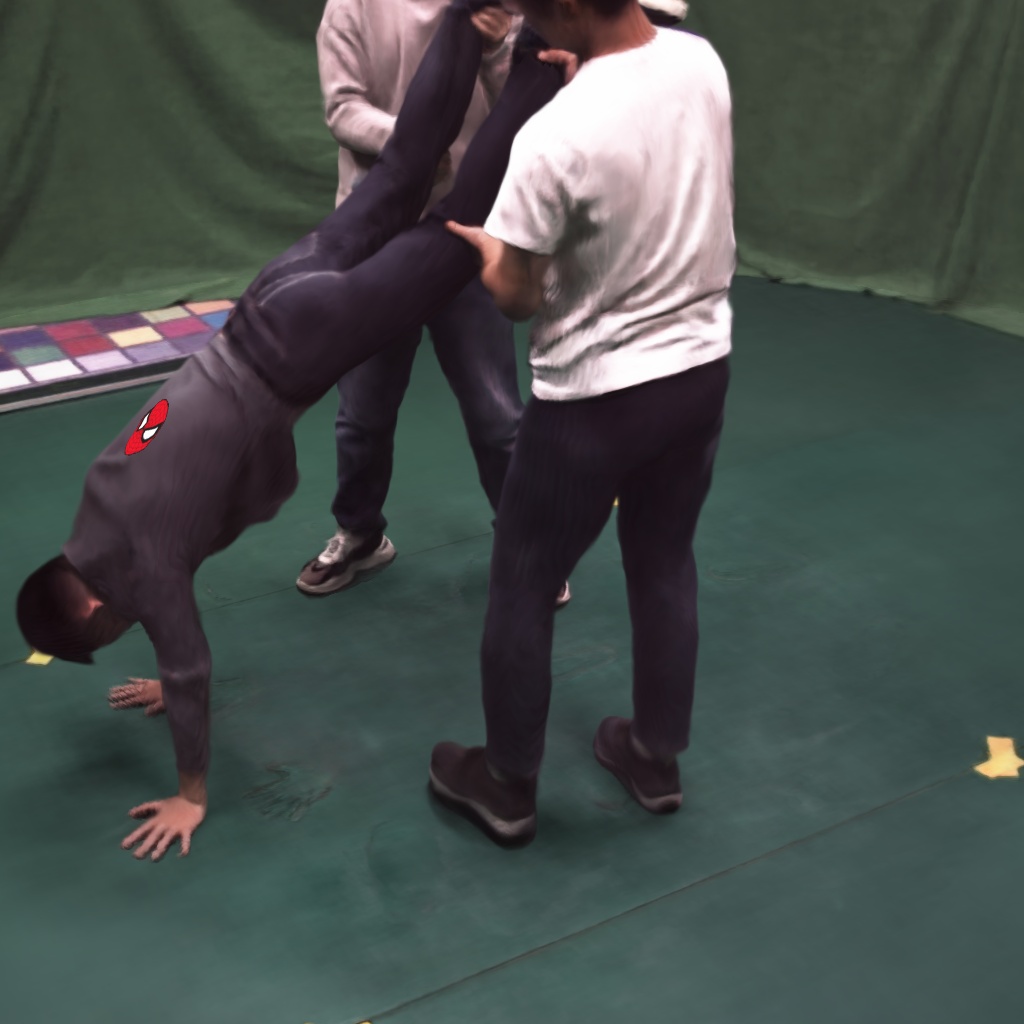}
& 
\includegraphics[width=\mywidth]{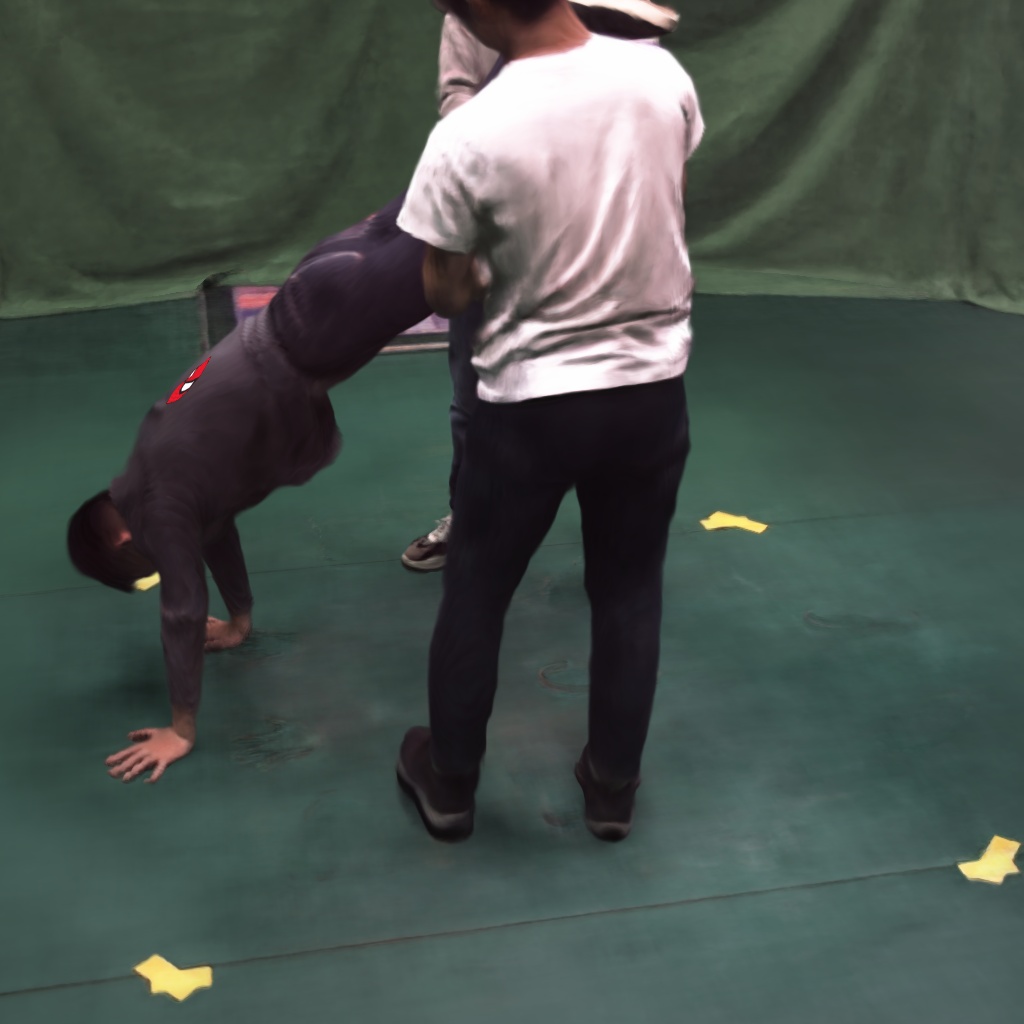}
&
\includegraphics[width=\mywidth]{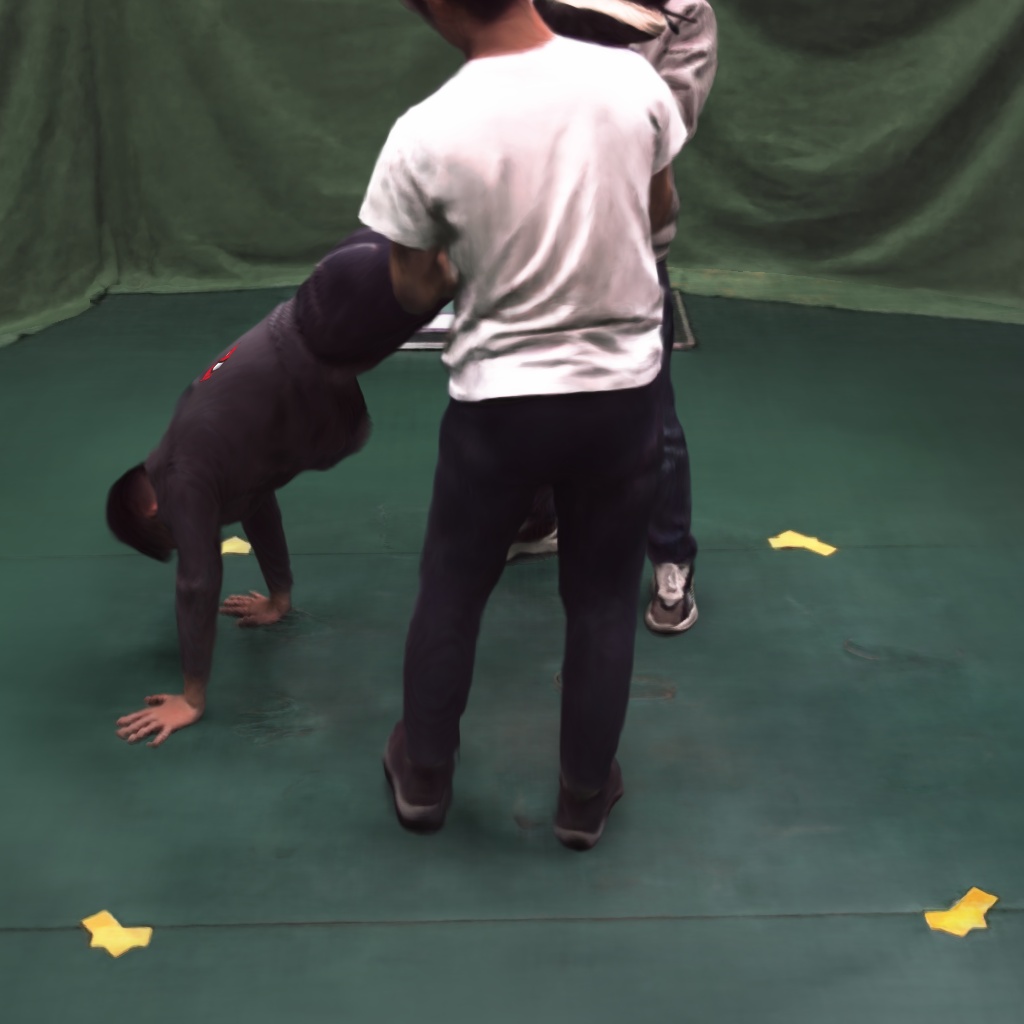}
\\
Frame 3
& Frame 4
& Frame 5
& Frame 6
\\[2ex]

\includegraphics[width=\mywidth]{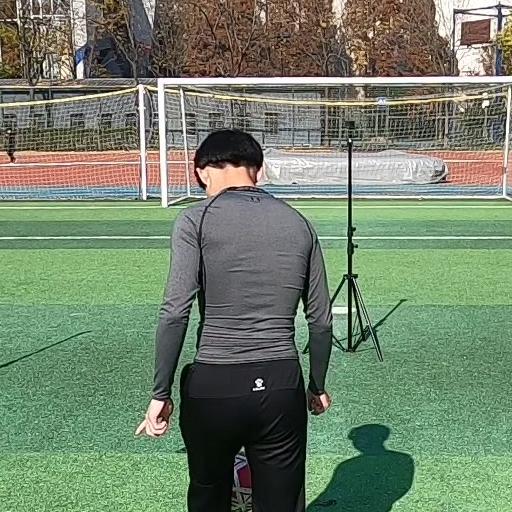}
&\includegraphics[width=0.1 \textwidth]{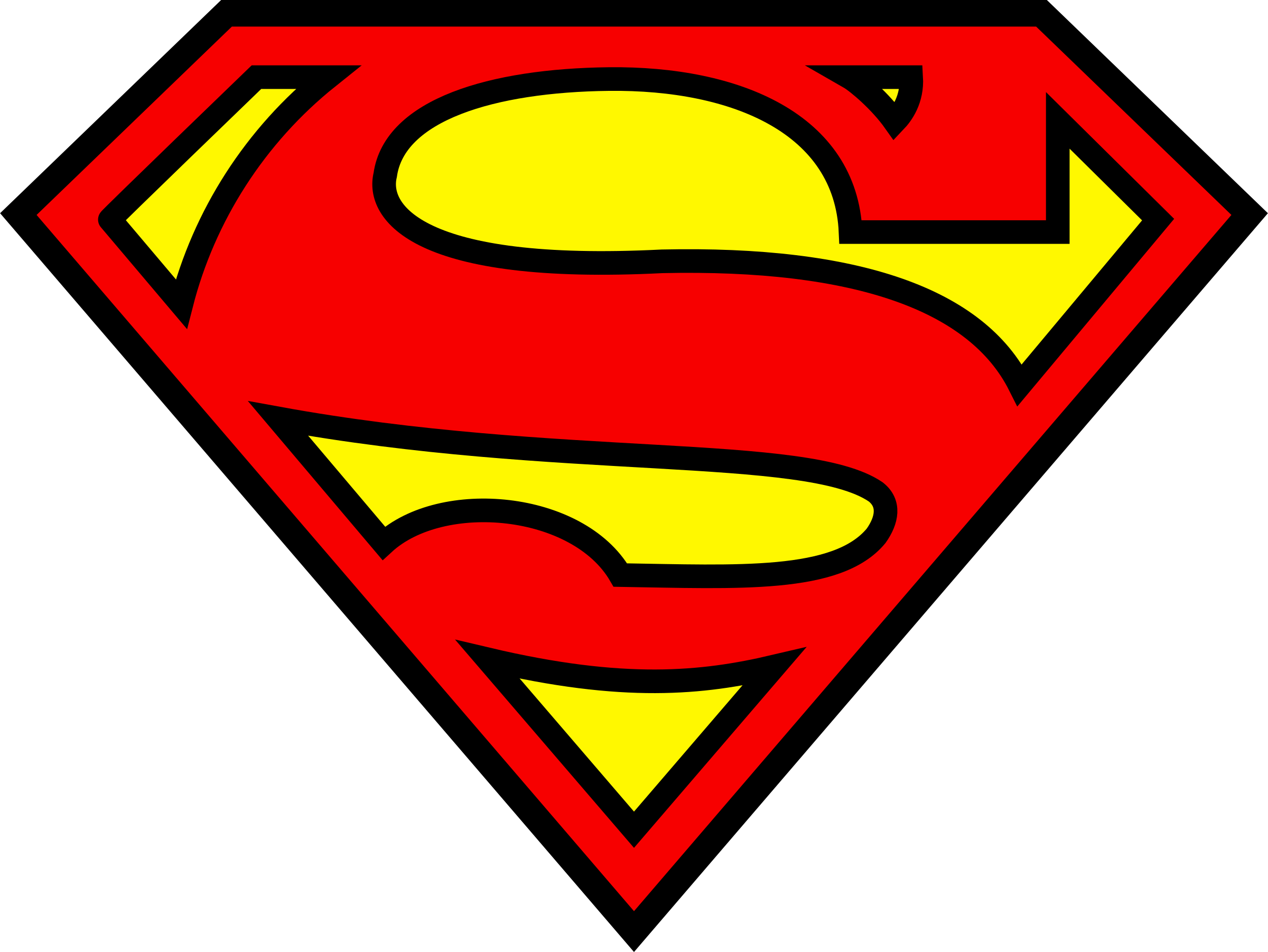}
&\includegraphics[width=\mywidth]{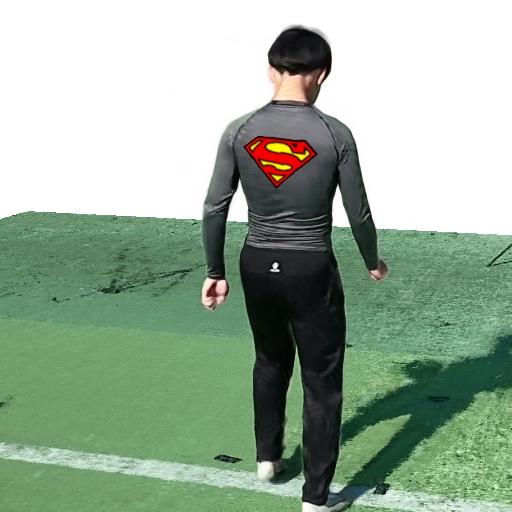}
&
\includegraphics[width=\mywidth]{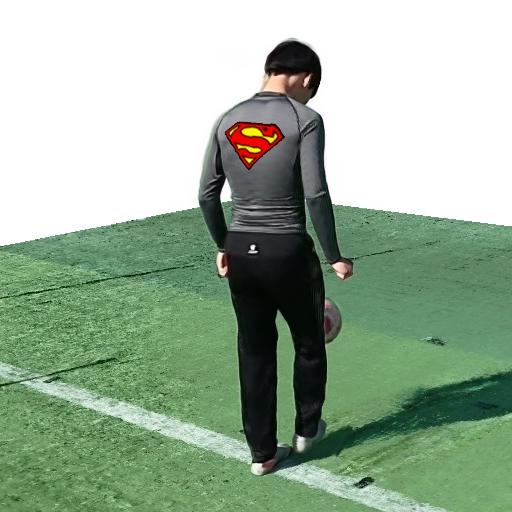}

\\
Original image
&
Added texture
&
Frame 1
& 
Frame 2

\\[2ex]
\includegraphics[width=\mywidth]{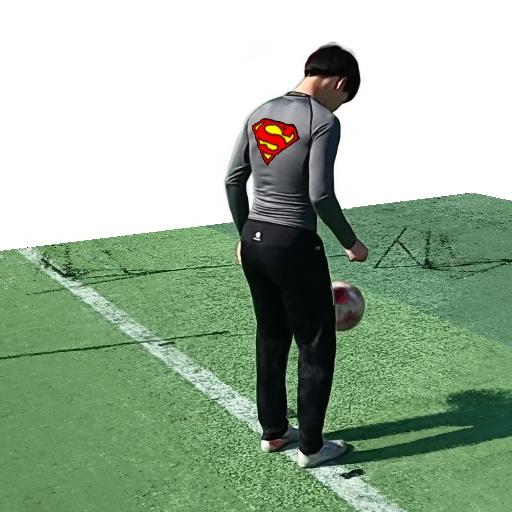}
&
\includegraphics[width=\mywidth]{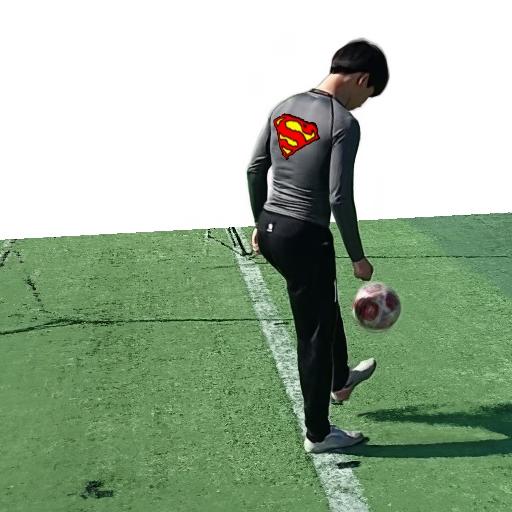}
& \includegraphics[width=\mywidth]{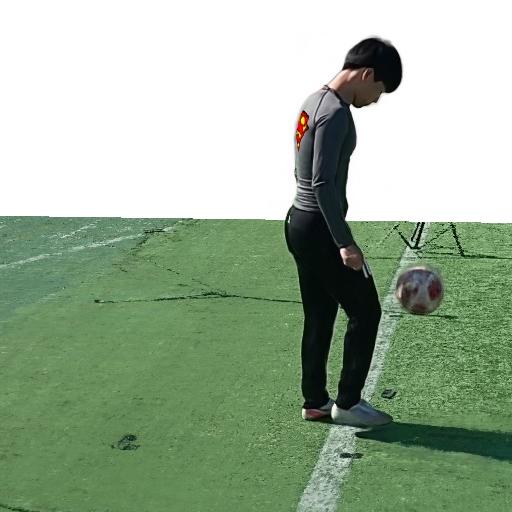}
& \includegraphics[width=\mywidth]{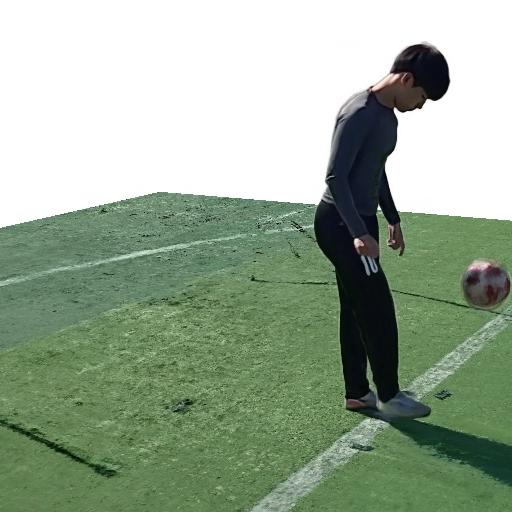}
\\

Frame 3
& Frame 4
& Frame 5
& Frame 6
\end{tabular}

\end{center}
\vspace{-0.2em}
\captionsetup{font={normalsize}} 
\caption{\textbf{Qualitative results on the ZJU-MoCap dataset.
} More results are shown in the supplementary video.
}
\vspace{-1em}

\label{zju}
\end{figure*}

\paragraph{Comparisons.}
We present some comparison results for our qualitative evaluation in~\figref{fig:exp}.
Our algorithm performs much better than other baselines.
Using optical flows and scene flows to warp the edited content is prone to accumulate errors, resulting in poor temporal consistency.
Additionally, because ``SF+DynNeRF'' lacks a well-defined and smooth surface, 
the rendered results of ``SF+DynNeRF'' and ``OF+DynNeRF'' at novel views suffer from blur and ghosting artifacts.
The results of Nerfies and HyperNeRF also exhibit serious temporal inconsistency due to inaccurate correspondences.
Additionally, HyperNeRF tends to overfit on a single edited frame, which seriously affects the results of other frames.
The user study results also show that our method is  preferred by users, as shown in~\tabref{usr}.
\paragraph{More qualitative results.}
In~\figref{nerfies}, we show the qualitative results on the Nerfies-HyperNeRF-CoNeRF dataset.
In~\figref{zju}, we show the qualitative results of the ZJU-MoCap dataset.
\subsection{Ablation Studies}
\begin{table}
    \centering
    \tabcolsep=0.2cm
    \caption{\textbf{Quantitative results of the ablation studies.} 
    We compare our method with baselines in terms of PCK-T and EPE. 
    }
    \begin{threeparttable}
    \begin{tabular}{c c c c c}
    \toprule
    & EPE$\downarrow$    & PCK-1$\uparrow$  & PCK-2$\uparrow$ \\
    \midrule
    Ours         & \textbf{0.72}    & \textbf{0.83}      &\textbf{ 0.99  }      \\
    Ours w/o Lap       & 1.33             & 0.35               & 0.81       \\
    Ours w/o FP     & 1.29             & 0.46              & 0.80         \\
    Ours w/o SFF     & 1.55 & 0.33   & 0.74 \\
    Ours w/o Inv &1.57 &0.37 & 0.65  \\
    SF warping & 1.34             & 0.34              & 0.88          \\
    \bottomrule
    \end{tabular}
    \end{threeparttable}
    \label{ab_corr}
    \vspace{-0.2in}
\end{table}


\paragraph{Temporal consistency.}

We analyze how the proposed components in motion representation affect temporal consistency.
We measure temporal consistency by evaluating pixel-wise correspondence accuracy between the reference image and other frames.
%
We adopt PCK-T~\cite{truong2021learning} and EPE~\cite{ilg2017flownet} to measure the pixel-wise correspondence accuracy.
In the context of PCK-T, the letter ``T'' represents a given pixel threshold, which we set to 1 and 2 in our experiments.
The ``Truck'' scene in the Nvidia Dynamic Scenes Dataset is selected for this experiment because it contains large areas of textureless planar regions, 
which is easy to obtain the ground truth of dense correspondence and is more challenging.
By annotating the four corners of the truck plane in each frame, the homography matrix is able to be calculated, and the ground truth of dense correspondence can be obtained.

We remove two important regularizations to verify their effectiveness, including the feature-based photo-metric term and the Laplacian smooth term.
The ablation experiments of removing the feature-based photo-metric term and Laplacian smooth term are abbreviated as ``Ours w/o FP'' and ``Ours w/o Lap'', respectively.
To demonstrate that we need scene flow fields to distill the motion information to invertible networks, we design a baseline named ``Ours w/o SFF''.
This baseline removes the scene flow fields. 
To supervise the invertible networks, we directly project the local surface to a 2D plane and then utilize the optical flow estimated by RAFT to constrain the position of the local surface between adjacent frames.
In order to show the effectiveness of invertible networks, we propose a baseline abbreviated as ``Ours w/o Inv''.
In this baseline, the invertible networks are replaced by forward-warping MLP and backward-warping MLP, like Disentangled3d~\cite{tewari2022disentangled3d}.
We design a baseline named ``SF warping'' which directly warps the edited region with scene flows to show that there will be cumulative errors.

As shown in~\tabref{ab_corr},  ``Ours w/o FP'' and ``Ours w/o Lap'' have lower PCK and higher EPE than our method,
which indicates that the regularization added on the local surface can improve the temporal consistency.
The performance of ``Ours w/o SFF'' demonstrates that it is necessary to utilize scene flow fields to distill the motion information to invertible networks.
This is because using scene flow fields to distill the motion information is a more direct 3D supervision signal than 2D optical flow  constraints.
Additionally, following NSFF~\cite{li2020neural} and Gao~\etal~\cite{Gao2021DynNeRF}, temporal photo-metric consistency and some regularization can be added to help eliminate noise in scene flow fields during training.
Because our method strictly satisfies cycle consistency, it is less likely to accumulate errors and performs better than  ``Ours w/o Inv''.
Our method is better than ``SF warping'',
for we add the regularization on the local surface to eliminate the accumulated errors.

\paragraph{Importance of handling occlusion relationship.}
\begin{figure}
\begin{center}
\setlength\tabcolsep{0.2em}
\newcommand{\mywidth}{0.2 \textwidth}
\begin{tabular}{cc}
\includegraphics[width=\mywidth]{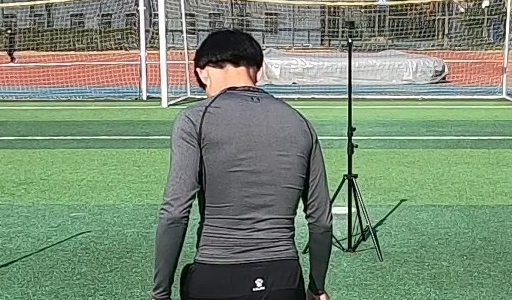}
&\includegraphics[width=\mywidth]{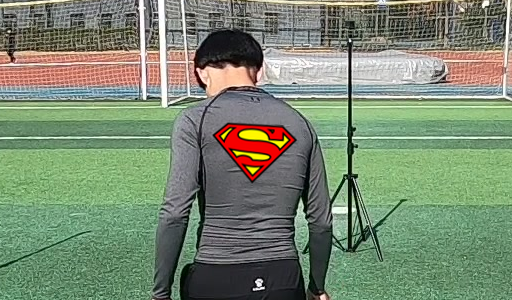}
\\
Original image
&
Edited image
\\
\includegraphics[width=\mywidth]{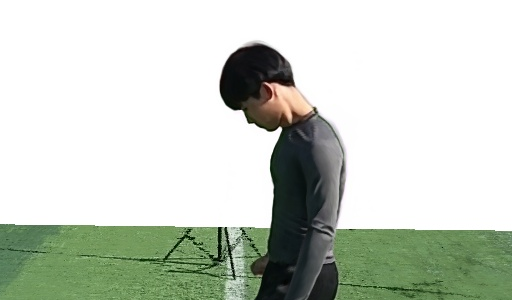}
&\includegraphics[width=\mywidth]{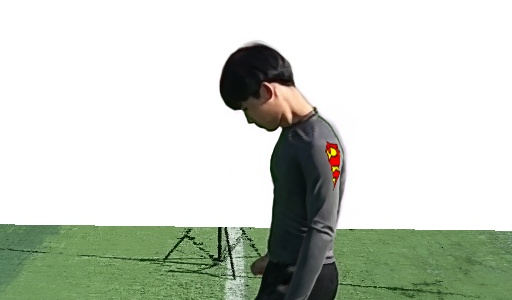}
\\
Ours
&
Ours w/o Occ
\end{tabular}
\end{center}
\vspace{-1em}
\captionsetup{font={normalsize}} 
\caption{\textbf{Importance of handling occlusion relationship.}
We show the results of our method with and without handling the occlusion relationship.
The baseline ``Ours w/o Occ'' fails to correctly handle the occlusion, resulting in the edited content behind the human body being visible. 
}
\vspace{-1em}

\label{occ}
\end{figure}

Our method combines the local surface and the given dynamic NeRF
and uses the volume rendering equation to render them to handle the occlusion relationship.
To demonstrate the significance of this design,
we directly render the local surface with the mesh renderer
and combine it with the rendered results of the dynamic NeRF.
We abbreviate this baseline as ``Ours w/o Occ''.
As shown in~\figref{occ}, if the occlusion relationship is not handled, the rendered results would be incorrect.
\section{Conclusions}
Editing the appearance of dynamic 3D scenes is an important and challenging task.
In this paper, we propose a method for editing the local appearance of dynamic 3D scenes by modifying individual frames in the video.
Through our proposed trackable local surface, we are able to precisely edit the local appearance of the dynamic scene in a temporally coherent manner. Experiments have demonstrated that our method can achieve high-quality results on a variety of dynamic scenes.


\bibliographystyle{ACM-Reference-Format}
\bibliography{ref}

\end{document}